\documentclass[a4paper,12pt]{article}
\usepackage[top=2.5cm, bottom=2.5cm, left=2.5cm, right=2.5cm]{geometry}
\usepackage{amsmath}
\usepackage{amssymb}
\usepackage{graphicx}
\usepackage{authblk}
\usepackage{bbm}
\usepackage{float}
\usepackage{chngcntr}
\usepackage[hidelinks]{hyperref}
\usepackage[authoryear,round]{natbib}
\usepackage[T1]{fontenc}

\DeclareMathOperator*{\argmax}{argmax}
\DeclareMathOperator*{\softmax}{softmax}
\DeclareMathOperator*{\OneHot}{OneHot}
\DeclareMathOperator*{\Encoder}{Encoder}

\DeclareMathOperator*{\Sum}{Sum}

\title{MEG-GPT: A transformer-based foundation model for magnetoencephalography data}

\author[1,2,*]{\small Rukuang Huang}
\author[1,3]{\small SungJun Cho}
\author[1,2]{\small Chetan Gohil}
\author[1,4]{\small Oiwi Parker Jones}
\author[1,2]{\small Mark Woolrich}

\affil[1]{\small Oxford Centre for Integrative Neuroimaging (OxCIN), University of Oxford, United Kingdom}
\affil[2]{\small Department of Psychiatry, University of Oxford, United Kingdom}
\affil[3]{\small Nuffield Department of Clinical Neurosciences, University of Oxford, United Kingdom}
\affil[4]{\small Department of Engineering Science, University of Oxford, United Kingdom}
\affil[*]{Corresponding author: rukuang.huang@psych.ox.ac.uk}

\date{}

\begin{document}
\maketitle	

\begin{abstract}
Modelling the complex spatiotemporal patterns of large-scale brain dynamics is crucial for neuroscience, but traditional methods fail to capture the rich structure in modalities such as magnetoencephalography (MEG). Recent advances in deep learning have enabled significant progress in other domains, such as language and vision, by using \textit{foundation models} at scale. Here, we introduce \texttt{MEG-GPT}, a transformer-based foundation model that uses time-attention and next time-point prediction. To facilitate this, we also introduce a novel data-driven \textit{tokeniser} for continuous MEG data, which preserves the high temporal resolution of continuous MEG signals without lossy transformations. We trained \texttt{MEG-GPT} on tokenised brain region time-courses extracted from a large-scale MEG dataset (N=612, eyes-closed rest, Cam-CAN data), and show that the learnt model can generate data with realistic spatio-spectral properties, including transient events and population variability. Critically, it performs well in downstream \textit{decoding} tasks, improving downstream supervised prediction task, showing improved zero-shot generalisation across sessions (improving accuracy from 0.54 to 0.59) and subjects (improving accuracy from 0.41 to 0.49) compared to a baseline methods. Furthermore, we show the model can be efficiently \textit{fine-tuned} on a smaller labelled dataset to boost performance in cross-subject decoding scenarios. This work establishes a powerful foundation model for electrophysiological data, paving the way for applications in computational neuroscience and neural decoding.
\end{abstract}


\vspace{0.5cm}

\section*{Highlights}
\begin{itemize}
\item \texttt{MEG-GPT}, a transformer-based foundation model that learns complex spatiotemporal dynamics from large-scale resting-state MEG data.
\item A novel data-driven \textit{tokeniser} converts continuous MEG signals into discrete tokens without loss of temporal or spectral resolution.
\item \texttt{MEG-GPT} generates realistic synthetic MEG signals that replicate key spatio-spectral properties, including transient events, inter-subject variability and subject-specific fingerprints.
\item Boosts \textit{zero-shot decoding accuracy} across subjects (from 0.41 to 0.49) and sessions (from 0.54 to 0.59) compared to traditional baselines.
\item The pre-trained model \textit{efficiently fine-tunes} on small, labelled datasets to enhance cross-subject decoding.
\end{itemize}

\section{Introduction} \label{sec:introduction}

A central aim in neuroscience is to understand the rich, dynamic patterns observed in large-scale brain activity. Electrophysiological imaging techniques provide a direct measure of neural activity with millisecond temporal resolution, making them particularly useful for studying fast brain dynamics. Among these techniques, magnetoencephalography (MEG) offers a non-invasive measure of neural activity that is  localised with good spatial accuracy and excellent temporal resolution~\citep{proudfoot2014magnetoencephalography}.

Traditional analyses of source-localised MEG data often involve averaging over space or time, leading to a loss of spatiotemporal information~\citep{tzovara2012decoding, brookes2010investigating, konig2015averaging, alexander2013traveling}. For instance, power spectral density (PSD) analyses  average over time and sometimes space~\citep{luck2014introduction}, while static functional connectivity analyses average over time~\citep{gohil2024dynamic}. Furthermore, these analyses are often applied to small datasets in isolation. The recent availability of large-scale MEG datasets, like Cam-CAN~\citep{taylor2017cambridge, shafto2014cambridge}, offers a new opportunity to move beyond these limitations. Modern deep learning approaches can be trained on large-scale data to extract rich spatiotemporal structure and patterns of population variability, which can then be leveraged to benefit inferences on smaller, specialized datasets~\citep{geron2022hands, murphy2012machine}.

\textit{Foundation models}~\citep{bommasani2021opportunities} are a class of models trained on large amounts of data at scale, often using \textit{self-supervised learning}~\citep{geron2022hands, murphy2012machine}. This contrasts with \textit{supervised learning}, which aims to learn a mapping from the data to \textit{labels}\footnote{Auxiliary quantities that we would like to predict.}. In self-supervised learning, the objective is to model the statistical dependencies within the data using a label derived from the data itself. In time series data, and in the approach we take here, this is often achieved by predicting the next time step; note that this can also be thought of as fitting an autoregressive generative model to the data. Foundation models can extract `general features' from the training data that can be applied to new datasets. Often, a pre-trained foundation model is \textit{fine tuned}\footnote{Trained for a short period. This can be with the same or a different objective (loss function).} on an independent dataset, which may be labelled, for a particular \textit{downstream} application, such as a prediction task~\citep{bommasani2021opportunities}. Adopting this approach has achieved remarkable success in modelling language and vision~\citep{radford2021learning, ramesh2021zero, brown2020language, touvron2023llama, dosovitskiy2020image, zhai2022scaling, nie2022time}. 

This approach is potentially valuable in the analysis of MEG data, where unlabelled resting-state data is abundant and labelled data, such as that for cognitive tasks or clinical populations, is limited. Improving supervised learning performance on MEG data is critical for a variety of neuroscience studies and applications. This includes: predicting a cognitive state based on data, referred to as \textit{decoding}~\citep{thomas2022interpreting}; disease classification and biomarker discovery~\citep{craik2019deep}; patient stratification \citep{bosl2018eeg,cassani2018systematic}; and brain-computer interfaces~\citep{lotte2018review, dash2019decoding, tang2023semantic}. Foundation models offer a powerful framework for such applications, as they can be adapted to specific tasks with minimal labelled data while leveraging prior knowledge from large-scale unlabelled data~\citep{bommasani2021opportunities}.

While existing foundation models for electrophysiological data often rely on time-frequency transformations, such as wavelet methods~\citep{wang2023brainbert, yuan2024brainwave}, that compromise data resolution, our approach circumvents this entirely. Inspired by the Generative Pre-Trained Transformer (GPT) family of models~\citep{radford2018improving, radford2019language, brown2020language}, we introduce \texttt{MEG-GPT}, a foundation model that learns directly from tokenised MEG data. \texttt{MEG-GPT} is a nonlinear autoregressive (AR) model~\citep{marple2019digital} based on a transformer architecture\footnote{More specifically, \texttt{MEG-GPT} is based on the decoder block of a transformer architecture.}~\citep{vaswani2017attention, geron2022hands}. This model predicts the next token from a sequence of previous tokens. This is made possible by another contribution: a bespoke, data-driven tokeniser for continuous electrophysiological data that operates with no loss in temporal or spectral resolution.

Foundation models are typically evaluated in terms of their performance on downstream tasks, such as classification or prediction~\citep{bommasani2021opportunities, zhai2022scaling}. However, the generative nature of GPT models means they are able to produce new synthetic data. This offers a complementary approach for evaluating model performance. We can assess whether the model can generate important features of interest that are present in the training data in the new synthetic data. In language and vision models, this is done by qualitatively examining the generated text~\citep{brown2020language, touvron2023llama} or images~\citep{ramesh2021zero}. Here, we evaluate \texttt{MEG-GPT} quantitatively by assessing the model's ability to generate realistic features, such as the transient spatio-spectral structure and inter-subject variability in neural data.

In the following, we first validate the performance of our tokeniser. Then, we train \texttt{MEG-GPT} on the tokenised parcel time courses extracted from the eyes closed resting-state data in Cam-CAN (612 subjects, $\sim2.8$ billion time points). Next, we demonstrate that the trained \texttt{MEG-GPT} can generate new data with realistic spectral properties, transient dynamics, and inter-subject variability. Finally, we showcase its practical utility in a downstream decoding task~\citep{king2014characterizing, grootswagers2017decoding}, where it achieves superior zero-shot generalisation compared to traditional approaches.


\section{Methods} \label{sec:methods}

\subsection{Tokeniser} \label{sec:methods/tokeniser}
GPT models typically learn statistical dependencies between \textit{tokens} of data. These are discrete `building blocks' of the data. In large language models~\citep{tunstall2022natural}, these tokens correspond to words (or individual characters). Here, we have continuous MEG data. Hence, before we can feed the data into the GPT model, we need a ``tokeniser'' that can map each time-series of continuous MEG data to a time series of discrete tokens; this process is then repeated separately for all channels or brain regions. Our proposed tokeniser accomplishes this using an autoencoder framework, which learns a discrete set of short, reusable temporal patterns -- the `tokens' -- directly from the data.

Training the foundation model on tokenised, rather than continuous, MEG data means we can benefit from using the cross entropy as the loss function (see Section~\ref{sec:methods/foundation_model/loss_function}). Cross entropy has been shown to have better convergence properties compared to the mean-square-error (MSE) loss, which is used for continuous data~\citep{geron2022hands}.

The simplest type of tokenisation would quantise the data into bins, e.g. after a $\mu$-law transformation~\citep{recommendation1988pulse, van2016wavenet, csaky2024foundational}. However, this does not account for the temporal-spectral properties of the data. We therefore took a data-driven approach. 

One option is the vector-quantised variational autoencoder (VQ-VAE)~\citep{van2017neural}, which learns to represent the data as a time series of discrete vectors with the data at each time point \textbf{compressed} into one of a set of vectors. The set of vectors is learnt from the data and is referred to as a \textit{codebook} (or \textit{dictionary}).

In contrast to this, we wanted a tokeniser that is not aiming to compress the data, i.e. we want it to be a lossless transform. This means that it does no regularisation, and has no loss of temporal or spatial resolution. Instead, we want all the modelling of rich spatio-temporal structure to be done downstream by the more sophisticated transformer-based foundation model.

\begin{figure}[!h]
    \centering
    \includegraphics[width=0.7\linewidth]{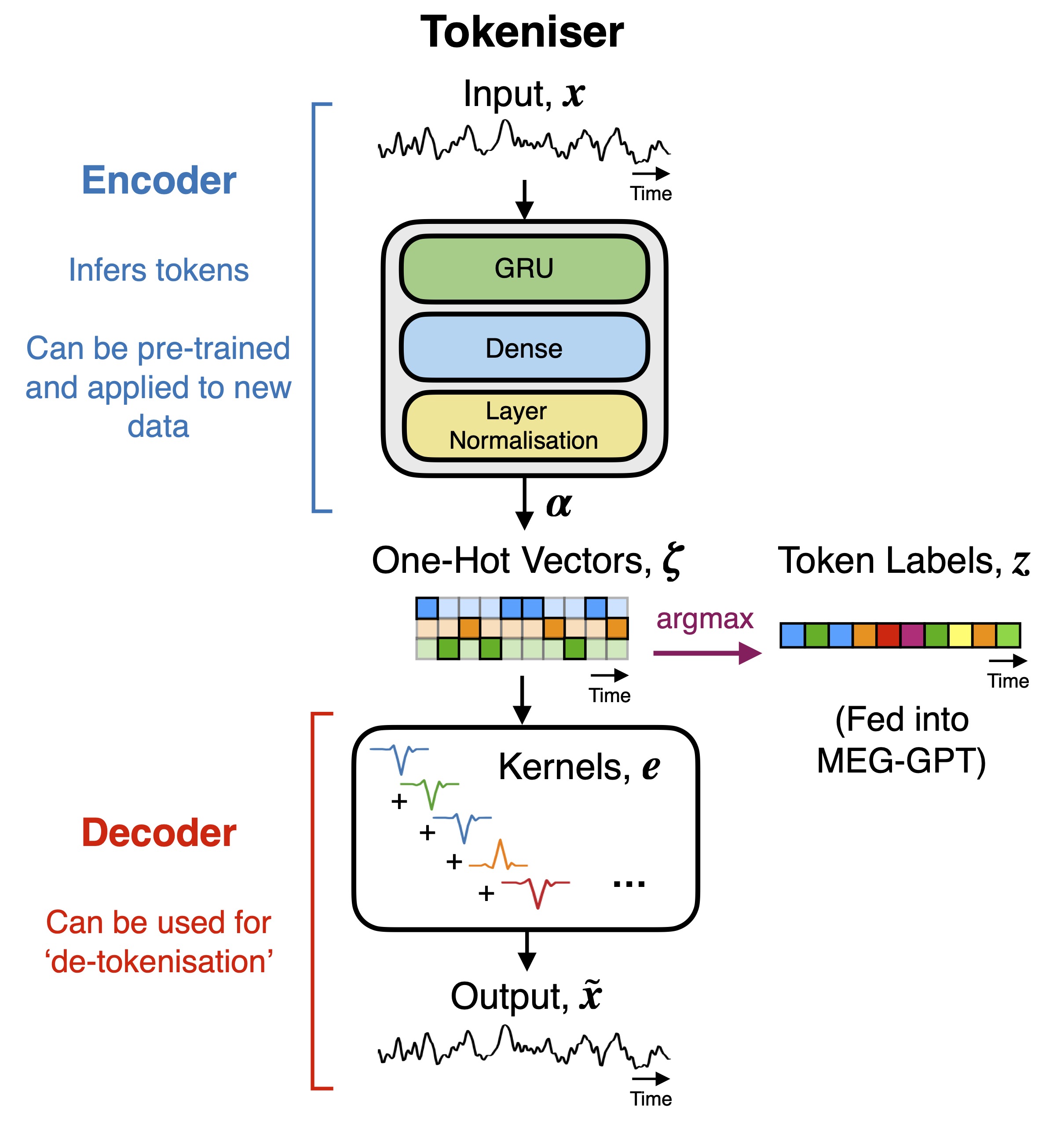}
    \caption{\textbf{Illustration of the tokeniser}. The input to the tokeniser is a single sequence of continuous MEG data and its encoder's output is a single sequence of token labels. The tokeniser is based on an autoencoder framework. The encoder maps the continuous data $\boldsymbol{x}$ onto logits $\boldsymbol{\alpha}$ that are used to calculate token labels $\boldsymbol{z}$. The decoder reconstructs the data $\tilde{\boldsymbol{x}}$ from token labels $\boldsymbol{z}$ using a weighted sum of token kernels $\boldsymbol{e}$. The token labels $\boldsymbol{z}$ are then fed into \texttt{MEG-GPT}.}
    \label{fig:tokeniser}
\end{figure}

Our tokeniser, illustrated in Figure~\ref{fig:tokeniser}, is described below. It is inspired by the VQ-VAE, but the codebook is embedded in the decoder. Note that, the tokeniser does not need a straight-through estimator (which is used in a VQ-VAE) to calculate the gradient, and that it is a standard autoencoder, i.e. no stochastic sampling is performed.

\subsubsection{Encoder} \label{sec:methods/tokeniser/encoder}

The \textit{encoder} learns a mapping from the continuous MEG data to sequence of categorical \textit{token labels}, which are unique indices for each token. The same tokeniser is applied to each channel independently.

Let $\boldsymbol{x} = [x_1, ..., x_T] \in \mathbb{R}^T$ be the continuous MEG data for a single channel, where $T$ is the number of time points. The goal of the encoder is to learn a sequence of categorical token labels $\boldsymbol{z} = [ z_1, ..., z_T ] \in \mathbb{R}^T$. The token label at each time point can take one of $K$ values (a pre-specified hyperparameter). The encoder learns a set of \textit{logits} at each time point $\boldsymbol{\alpha}_t = [\alpha_1, ..., \alpha_K] \in \mathbb{R}^K$, which reflects the underlying probability of each token:
\begin{equation}
\boldsymbol{\alpha}_t = \Encoder ( \boldsymbol{x} ).
\end{equation}
In our tokeniser, the encoder is a single layer of the GRU (Gated Recurrent Unit)~\citep{cho2014learning} layer and a Dense layer, followed by a Layer Normalisation~\citep{ba2016layer}. The token label is calculated from the logits using
\begin{equation}
\label{eq:tokeniser-z}
z_t = \argmax ( \boldsymbol{\alpha}_t ).
\end{equation}

\subsubsection{Decoder} \label{sec:methods/tokeniser/decoder}

The continuous MEG data is built by combining the tokens (i.e. building blocks). The decoder reconstructs the continuous data based on the token labels inferred by the encoder. To model the temporal characteristics of the data, we decided to use a dictionary of tokens based on 1D convolution kernels. These kernels are learnt from the data alongside the token labels.

Let $\boldsymbol{e} = [ \boldsymbol{e}^1, ..., \boldsymbol{e}^K] \in \mathbb{R}^{K \times d_\mathrm{token}}$ be the set of \textit{token kernels}, where each kernel is a vector of dimensionality $d_\mathrm{token}$ (a pre-specified hyperparameter). We reconstruct the data at each time point as a weighted sum of token kernels:
\begin{equation}
    \tilde{\boldsymbol{x}}_t = \sum_{\tau=- d_\mathrm{token} / 2}^{d_\mathrm{token} / 2} w_{\tau} e^{z_{t+\tau}}_{\tau},
\end{equation}
where $\boldsymbol{w} = [w_1, ..., w_{d_\mathrm{token}}] \in \mathbbm{R}^{d_\mathrm{token}}$ are learnable weights in the decoder.

\subsubsection{Training} \label{sec:methods/tokeniser/training}

The tokeniser is trained using \textit{stochastic gradient descent}~\citep{geron2022hands}, where the learnable parameters of the model are updated iteratively to minimise a \textit{loss function}. In this work, the parameter updates were calculated using an Adam optimiser~\citep{kingma2014adam}.

\subsubsection*{Loss function}

The loss function is the MSE between the reconstructed data $\tilde{\boldsymbol{x}}$ and the input data $\boldsymbol{x}$.

\subsubsection*{Annealing}

An important component of our tokeniser is the \textit{argmax} operation that maps the logits to a categorical token label (Equation~\eqref{eq:tokeniser-z}). A key challenge is that the argmax operation is non-differentiable, which prevents end-to-end training.\footnote{The argmax operation is not differentiable, which impedes the backpropagation algorithm, and prevents us from calculating the gradient for model parameters that precede this operation.} To overcome this, we use an annealing technique to replace the $\argmax$ operation with a weighted sum of $\argmax$ and $\text{softmax}$\footnote{The softmax operation can be performed incorporating a \textit{temperature} parameter, which we set to a small value (of 0.1) in this paper to encourage categorical-like outputs.} during training:
\begin{equation}
z_t = (1 - \kappa) \cdot \argmax (\boldsymbol{\alpha}_t) + \kappa \cdot \softmax(\boldsymbol{\alpha}_t),
\end{equation}
where $\kappa$ starts from 1 at the beginning of the training process and gradually decreases to 0. This allows end-to-end training to occur by smoothly transitioning from a \textit{softmax} to an \textit{argmax}-like output during training. After training (during inference), $\kappa$ is set to zero.

\subsubsection{Token re-factorisation} \label{sec:methods/tokeniser/token_refactorisation}

In practice, after we trained the tokeniser, we found that not all tokens were used to reconstruct the data. Because these tokens do not appear in the data, we do not need to include them when training the foundation model. We performed \textit{token re-factorisation} to remove these tokens. This involves:
\begin{enumerate}
\item Relabelling the $K$ tokens in descending order (i.e. from 1 to $K$) in terms of their rate of occurrence in the tokenised training data.
\item Identifying the tokens that do not appear in the tokenised training data
\item Assigning the tokens with zero occurrence a label of 0.
\end{enumerate}
Following token re-factorisation, we end up with $K^*$ tokens, where tokens with labels from 1 to $K^* - 1$ correspond to tokens that appeared in the training data of the tokeniser.

\begin{figure}[!t]
    \centering
    \includegraphics[width=\linewidth]{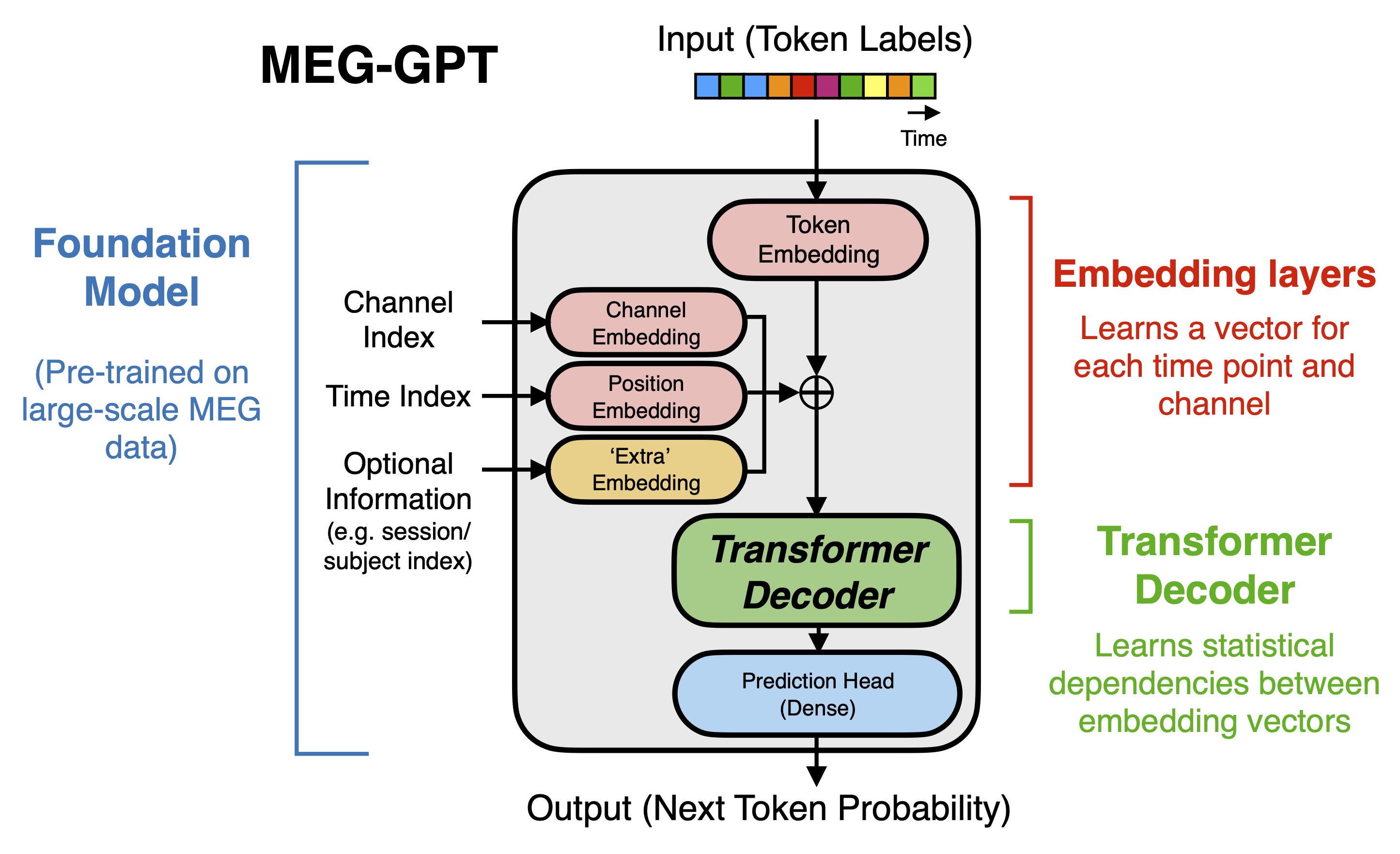}
    \caption{\textbf{MEG-GPT foundation model}. MEG-GPT is a nonlinear autoregressive model that predicts the token at the next timepoint from a preceding sequence of token labels. However, unlike classical autoregressive models, the autoregressive weighting can change as a function of the data. A key part of the foundation model is learning an \textit{embedding space} for the tokens, which generates vector for each token $\boldsymbol{v}_z$, channel $\boldsymbol{v}_c$, temporal position $\boldsymbol{v}_p$, any additional information $\boldsymbol{v}_s$ (e.g. subject ID) and combines (adds) these to provide a single embedding vector $\boldsymbol{v}$ at each time point. This combined embedding vector is fed into the Transformer Decoder, and the output of the Transformer Decoder is used to predict the next token via the Prediction Head.}
    \label{fig:meg-gpt}
\end{figure}

\subsection{The foundation model: MEG-GPT} \label{sec:methods/foundation_model}

\texttt{MEG-GPT} is a transformer-decoder-based foundational model that can be pre-trained on a large-scale datasets. The input to \texttt{MEG-GPT} is a sequence of token labels and the output is the probability for each token to be next. The architecture for \texttt{MEG-GPT} is shown in~Figure~\ref{fig:meg-gpt}. We describe each part in detail below.

\subsubsection{Input embedding} \label{sec:methods/foundation_model/input_embedding}

The model first transforms its discrete token label inputs into a rich, continuous vector space (the \textit{token embedding space}) through learnt embeddings. This process assigns a unique vector to each token label. This can be thought of as a compressed summary of characteristics of each token, e.g. tokens that have similar characteristics will get grouped together in the embedding space.

We also learn embedding vectors for the other inputs provided to \texttt{MEG-GPT}, i.e. the channel index, time index, and any other auxiliary information like a session, participant or task index. Each embedding vector is denoted by:
\begin{itemize}
\item $\boldsymbol{v}_z \in \mathbb{R}^{K^* \times d_z}$ for the token embedding.
\item $\boldsymbol{v}_c \in \mathbb{R}^{C \times d_c}$ for the channel embedding, where $C$ is the number of channels.
\item $\boldsymbol{v}_p \in \mathbb{R}^{L \times d_p}$ for position (time index) embedding, where $L$ is the sequence length.
\item $\boldsymbol{v}_s \in \mathbb{R}^{N \times d_s}$ for the additional information embedding, e.g. the participant ID, where $N$ is the number of unique identifiers for each piece of additional information.
\end{itemize}
$d_{*}$ denotes the length of each embedding vector. All of these embeddings are summed to obtain the overall input embedding
\begin{equation}
\boldsymbol{v} = \Sum (\boldsymbol{v}_z, \boldsymbol{v}_c, \boldsymbol{v}_p, \boldsymbol{v}_s) ~ \in \mathbbm{R}^{L\times C \times d},
\end{equation}
which is passed to the Transformer Decoder block. A Dense layer is used to map the individual embedding vector lengths ($d_z$, $d_c$, $d_p$, $d_s$) to $d$ if there is a mismatch in dimensionality. All embeddings are learnt when the \texttt{MEG-GPT} is trained. This combined vector, $\boldsymbol{v}$, effectively encodes the `\textbf{what}' (token), `\textbf{where}' (channel), `\textbf{when}' (time) and `\textbf{who}' (participant) of the signal, providing a comprehensive input to the Transformer Decoder block.

\subsubsection{Transformer and Prediction Head} \label{sec:methods/foundation_model/decoder_and_prediction_head}

The input to the Transformer Decoder is $\boldsymbol{v} \in \mathbbm{R}^{L\times C \times d}$, which is a time ($L$) by channels ($C$) by embedding length ($d$) tensor. The Transformer Decoder models the statistical dependencies between these embedding vectors. It can be understood as a powerful nonlinear autoregressive (AR) model, where the AR coefficients are not fixed but instead adapt dynamically to the input data via its multi-head attention mechanism.

\begin{figure}[H]
    \centering
    \includegraphics[width=\linewidth]{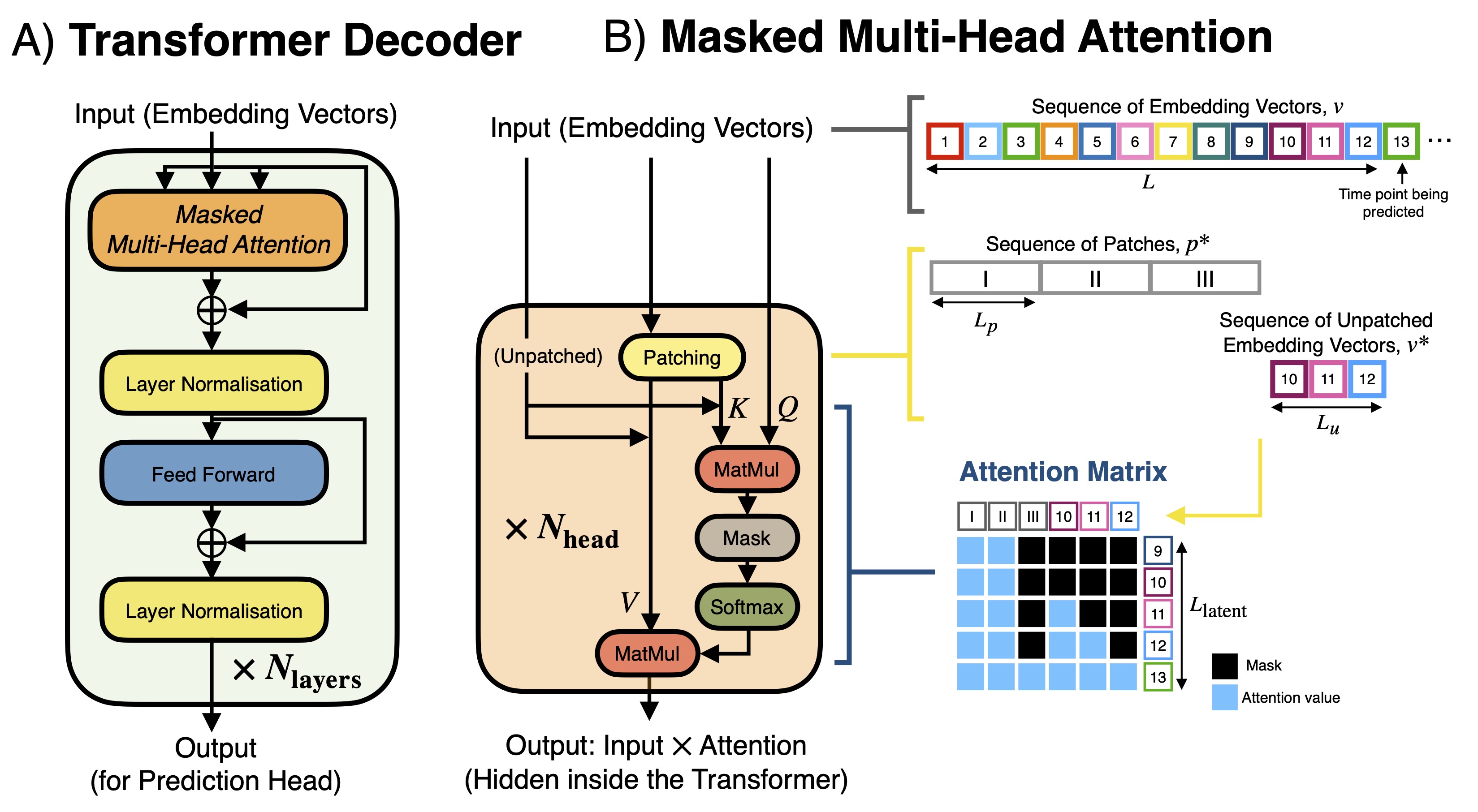}
    \caption{\textbf{Transformer Decoder and the masked multi-head attention used in MEG-GPT}. A) Architecture of the Transformer Decoder in Figure~\ref{fig:meg-gpt}. B) Masked multi-head (self-)attention used in the Transformer Decoder. Shows an example of predicting the token at the next time step with a receptive field of $L = 12$. The input embeddings $\boldsymbol{v}$ are divided into 3 patches of patch size $L_p = 4$, resulting in \textit{patched inputs} $\boldsymbol{p}^*$. The last $L_u = 3$ unpatched embeddings are retained as the \textit{unpatched inputs} $\boldsymbol{v}^*$, allowing for finer temporal resolution near the prediction point. The attention matrix used in the self-attention mechanism has a latent sequence length of $L_\mathrm{latent} = 5$ corresponding to the output tokens. Each output attends to both patched inputs $\boldsymbol{p}^*$ and unpatched inputs $\boldsymbol{v}^*$. A temporal mask (indicated by black squares) is applied to prevent information leakage from future time points during training.}
    \label{fig:transformer}
\end{figure}

The Transformer Decoder is based on the architecture in~\citep{vaswani2017attention}. Figure~\ref{fig:transformer}A shows the architecture of the \texttt{MEG-GPT} Transformer Decoder. In each layer, we use residual connections around the Masked Multi-Head Attention block and a Feed Forward layer, followed by Layer Normalisation. The output of the Transformer Decoder is passed to the Prediction Head -- a single Dense layer -- which generates the final logits used to predict the probability of the next token.

\subsubsection{Masked multi-head attention} \label{sec:methods/foundation_model/multi_head_attention}

A key innovation in the Transformer Decoder~\citep{vaswani2017attention} is the use of \textit{self-attention} layers. Here, the Transformer Decoder can learn to \textit{attend} to different parts of the input data's history in different ways that depend on the input data. How far back in time the Transformer Decoder can attend depends on the \textit{receptive field}. A key challenge for transformers is that the computational cost of the self-attention mechanism scales quadratically with sequence length, making it prohibitive for long time-series. To extend \texttt{MEG-GPT}'s receptive field without incurring this significant computational cost, we incorporate several recent modifications into the standard self-attention architecture, as illustrated in Figure~\ref{fig:transformer}B. These include patching, the use of unpatched sequences, and the Perceiver AR architecture.

\textbf{Patching}. Patching is a technique to divide the input embeddings $\boldsymbol{v} \in \mathbbm{R}^{L\times C\times d}$ into non-overlapping patches of length $L_p$, called the patch size. The resulting patched input embeddings becomes $\boldsymbol{p} \in \mathbbm{R}^{P\times L_p\times C\times d}$. As discussed in \citep{nie2022time}, aggregating time steps within patches can help extract local semantic information, which is typically not available in the single time level, and reduce computational cost. A linear Dense layer (shared across all patches) is used to collapse the $L_p$ dimension and extract information at the patch level, resulting in patched input $\boldsymbol{p}^* \in \mathbbm{R}^{P\times C\times d}$.

\textbf{Unpatched sequences}. Although patching can extract local information, it does mean there is a loss of temporal resolution in the time attention. This could be particularly detrimental to the model's ability to predict tokens over the short time scales, e.g. over time scales that correspond to the temporal window width of the token kernels ($d_\mathrm{token}$). Hence, in addition to the patched inputs, unpatched inputs $\boldsymbol{v}^* = \boldsymbol{v}[T - L_u + 1: T] \in \mathbbm{R}^{L_u\times C \times d}$ are also used to predict the future time point.

\textbf{Perceiver AR}. Our Transformer Decoder also makes use of the Perceiver AR architecture \citep{hawthorne2022general} to encode input sequences to latent representations of shorter sequence length $L_\mathrm{latent}$, and keep the autoregressive nature of the model by applying the correct masking. In this paper, we only shrink the sequence length of the latent representations in the first layer of the Transformer Decoder, so that all subsequent layers work with inputs of length $L_\mathrm{latent}$, reducing computational cost and memory requirement.

\subsubsection{Loss function} \label{sec:methods/foundation_model/loss_function}

\texttt{MEG-GPT} is trained using stochastic gradient descent using the Adam optimiser~\citep{kingma2014adam}. The loss function used in training is the cross entropy~\citep{murphy2012machine} between the predicted token probabilities and the real token labels. The loss for a sequence of token labels is computed as
\begin{equation}
\mathcal{L} = - \sum_{t=L - L_\mathrm{loss}}^L \sum_{k=1}^{K^*} \left\{ \OneHot (\boldsymbol{z}) \right\}_{tk} \log (\hat{p}_{tk}),
\end{equation}
where $\hat{p}_{tk}$ is the predicted probability of token $k$ at time $t$. Notice we only calculate the loss based on the last $L_\mathrm{loss}$ time points in the sequence. This is because the first few predictions made by \texttt{MEG-GPT} in the sequence have a relatively short input to base predictions on. Only including the predictions at the end of the sequence improves the stability of the computed loss.

\subsubsection{Generating new data} \label{sec:methods/generating_data}
Once trained, \texttt{MEG-GPT} can be used as a generative model to synthesise new MEG data by following a three-step process:

\begin{enumerate}
    \item \textbf{Prompt Initialisation}. A prompt sequence of tokens  $\boldsymbol{z}^* \in \mathbbm{R}^{L\times C}$ is created by sampling tokens from a categorical distribution weighted by their rate of occurrence in the training data. Extra inputs used during training (such as channel labels, session/participant indices) should also be provided.
    
    \item \textbf{Autoregressive Generation}. The model then predicts subsequent tokens one at a time. At each step, the next token is drawn from the model's output probability distribution using \textit{nucleus sampling} (top-p = 0.99)~\citep{holtzman2019curious}. This restricts sampling to the smallest possible set of tokens whose cumulative probability exceeds 0.99.

    \item \textbf{Signal Reconstruction}. The complete sequence of generated token labels is then passed through the pre-trained tokeniser's decoder. This converts the discrete sequence back into a continuous, synthetic MEG time series, reversing the tokenisation process.
    
\end{enumerate}

\subsection{Fine tuning the foundation model} \label{sec:methods/fine_tuning}

After the foundation model has been trained, it can then be \textit{fine-tuned} on another dataset to adapt to aspects that may not be present in the original training data. For example, in MEG data this may be potential variations in recording devices, sensor configurations, preprocessing pipelines, population, experimental design, etc. Fine tuning allows the model to adapt to the specific statistical and physiological characteristics of the new dataset, improving performance on downstream tasks. 

Fine-tuning process is computationally efficient. It typically involves training the model on the new, often smaller, dataset for only a few epochs using a small learning rate. This prevents the model from catastrophically forgetting the general features learnt during pre-training while gently adjusting weights to fit the new data.

A critical aspect of fine-tuning is determining which model components to update. Embeddings specific to the original training set, such as \textbf{subject embeddings}, must be discarded and retrained on the new dataset. In contrast, embeddings that are transferable, such as position (time) and channel (space) embedding, can be held fixed. Exact choices for the datasets studied in this work are given in Section~\ref{sec:methods/training_and_testing_datasets}.

\subsection{The foundation model as a feature extractor} \label{sec:methods/feature_extractor}
To evaluate \texttt{MEG-GPT}'s practical utility, we use it to extract features for a downstream decoding task. We compare the performance of a classifier when trained on three distinct feature sets derived from the same underlying MEG data epochs:

\begin{itemize}
    \item \textbf{Baseline Features}: Parcel MEG time courses are time-locked to the task onset and epoched. The epochs (with dimensions $C\times L$) are flattened and used as features.
    \item \textbf{Zero shot Features}: The prediction head of the \texttt{MEG-GPT} foundation model is discarded and the outputs (with dimensions $C\times L \times d$) from the decoder are collapsed over the time dimension by taking the average. The resulting outputs (with dimensions $C\times d$) are flattened and used as features.
    \item \textbf{Fine-tuned Features}: \texttt{MEG-GPT} is fine-tuned on the training set of the task dataset (see Section \ref{sec:methods/training_and_testing_datasets}) and features are extracted in the same way as the zero-shot case.
\end{itemize}

\subsection{Datasets} \label{sec:methods/datasets}

Two publicly available MEG datasets were used: one for training \texttt{MEG-GPT} (Cam-CAN) and another to illustrate a downstream decoding task (Wakeman-Henson). Both datasets were collected using an Elekta Neuromag Vectorview scanner at a sampling frequency of 1\,kHz and were processed in the same way.

\textbf{Cam-CAN}~\citep{taylor2017cambridge, shafto2014cambridge}. We used resting-state (eyes closed; $\sim$8.5 minutes) recordings from 612 healthy participants (310 males, 302 females, aged 18-88 years) from this dataset.

\textbf{Wakeman-Henson}~\citep{wakeman2015multi}. This dataset contains 19 healthy participants (11 males, 8 females, aged 23-37 years), which were scanned 6 times each. In each recording session, each participant performed a visual perception task. They were presented with 3 types of visual stimuli, including an image of a famous, unfamiliar or scrambled face. Each recording session was around 7.5 minutes and contain $\sim$200 trials which are evenly split across the three stimulus types. To ensure participants focus on the image, they were also asked to press one of the keys depending on whether they regarded each of the images as symmetric.

\subsubsection{Preprocessing} \label{sec:methods/datasets/preprocessing}

Both public MEG datasets were band-pass filtered between 0.03 and 330\,Hz and MaxFiltered~\citep{taulu2006spatiotemporal}. They were then further preprocessed using the \texttt{osl-ephys} toolbox, which is based on MNE~\citep{van2025osl}:
\begin{enumerate}
\item Band-pass filtered between 0.5 and 125\,Hz.
\item Notch filtered at 50 and 100\,Hz to remove power line artefacts.
\item Downsampled to 250\,Hz.
\item Automated bad segment and channel detection using the generalised extreme Studentised deviate procedure \citep{rosner1983percentage}.
\item FastICA \citep{hyvarinen1999fast} with 64 components to detect artefacts. Components with high correlation (threshold of 0.9) with the electrooculogram/electrocardiogram (EOG/ECG) channels were marked as noise and removed. Between 0 and 3 EOG components were rejected in each recording (mean 0.99, standard deviation 0.79) and between 0 and 5 ECG components were rejected (mean 2.25, standard deviation 0.84).
\end{enumerate}

\subsubsection{Coregistration and source reconstruction} \label{sec:methods/datasets/coregistration_and_source_reconstruction}

Coregistration and source reconstruction were also performed using the \texttt{osl-ephys} toolbox: coregistration was carried out using the osl-ephys tool RHINO, making use of a structural magnetic resonance imaging image and digitised headshape points (acquired with a Polhemus pen) for each subject; then, the data were source reconstructed onto an 8\,mm isotropic grid using a volumetric linearly constrained minimum variance  (unit noise gain) beamformer~\citep{van1988beamforming, van1997localization}.

\subsubsection{Parcellation, leakage correction and sign flipping} \label{sec:methods/datasets/parcellation_and_sign_flipping}

Parcellation, leakage correction and sign flipping was also performed with the \texttt{osl-ephys} toolbox. An anatomical parcellation was used to estimate the activity as 52 regions of interest using the first principal component across voxels associated with each parcel. Details of the parcellation can be found in~\citep{glasser52}.

A common problem in the estimation of source activity using electrophysiological data, is `spatial leakage' in the activity between neighbouring regions, which can lead to `ghost interactions' \citep{colclough2015symmetric}. The symmetric multivariate leakage reduction algorithm proposed by \citep{colclough2015symmetric} was used to reduce spatial leakage and ghost interactions, by removing all zero-lag correlation between parcel time courses.

The principal component analysis (PCA) step in performing the parcellation means the sign of each parcel time course is arbitrary. This poses a challenge for group-level analysis. We used the method proposed in \citep{vidaurre2016spectrally} to align the sign of each parcel time course across sessions/subjects. This algorithm uses  a greedy search based on randomly flipping the sign of each parcel time course to maximises the agreement between different sessions/subjects.

Finally, we temporally standardise each sign-flipped parcel time course (i.e. subtract the temporal mean and divide by the temporal standard deviation). All subsequent analysis is done on these standardised parcel (i.e. brain region) time courses.

\subsubsection{Training, validation and test sets} \label{sec:methods/training_and_testing_datasets}

We divide the Cam-CAN and the Wakeman-Henson datasets into training, validation and testing sets, which differ depending on the study:

\begin{itemize}
\item \textbf{Tokeniser}: The training set of the tokeniser includes the parcel time courses of the first 50 of 612 subjects in the Cam-CAN dataset. The testing set of the tokeniser includes the rest of the subjects in the Cam-CAN dataset and the entire Wakeman-Henson dataset. The hyperparameters for training the tokeniser are shown in Table~\ref{tab:tokeniser_hyperparameters}.

\item \textbf{MEG-GPT}: For training \texttt{MEG-GPT}, we employ a nine-to-one train-validation split for each subject in the Cam-CAN dataset, i.e. for each subject in the Cam-CAN dataset, 90\% of their data is used for training and 10\% of the data is used for evaluating the validation loss and accuracy. The hyperparameters for training \texttt{MEG-GPT} are available in Table~\ref{tab:foundation_hyperparameters_camcan}.

\item \textbf{Fine-tuning and task decoder}: During fine-tuning \texttt{MEG-GPT} and training of task decoder on the Wakemen-Henson dataset, the first 5 sessions of the first 18 subjects are used as the training set. The testing set includes the sixth session of the first 18 subjects (used for testing within subject generalisability) and all sessions of subject 19 (used for testing out of subject generalisability). When we fine-tuned the \texttt{MEG-GPT} model, we froze the token, channel and position embeddings, and only trained the Transformer Decoder and Prediction Head. The subject embeddings learnt from the Cam-CAN dataset were discarded at this stage. The hyperparameters for fine-tuning on the Wakeman-Henson dataset are available in Table~\ref{tab:foundation_hyperparameters_wh}.
\end{itemize}

\section{Results} \label{sec:results}

\subsection{The tokeniser reconstructs MEG data with high accuracy and generalises to unseen data} \label{sec:results/tokeniser_reconstructs}

First, we studied the performance of the novel data-adaptive tokeniser after training it on a subset of the Cam-CAN dataset (50 subjects), resulting in $K^* = 61$ tokens after token re-factorisation. Details regarding the hyperparameters of the tokeniser are given in Appendix~\ref{sec:extra_tokeniser_details}, along with the training curve (loss vs number of training epochs, Figure~\ref{fig:tokeniser_loss}) and distribution of token occurrences and token shapes (which shows fundamental building blocks of MEG data learnt by the model) of the final model (Figure~\ref{fig:token_summary}).

\begin{figure}[H]
    \centering
    \includegraphics[width=0.9\linewidth]{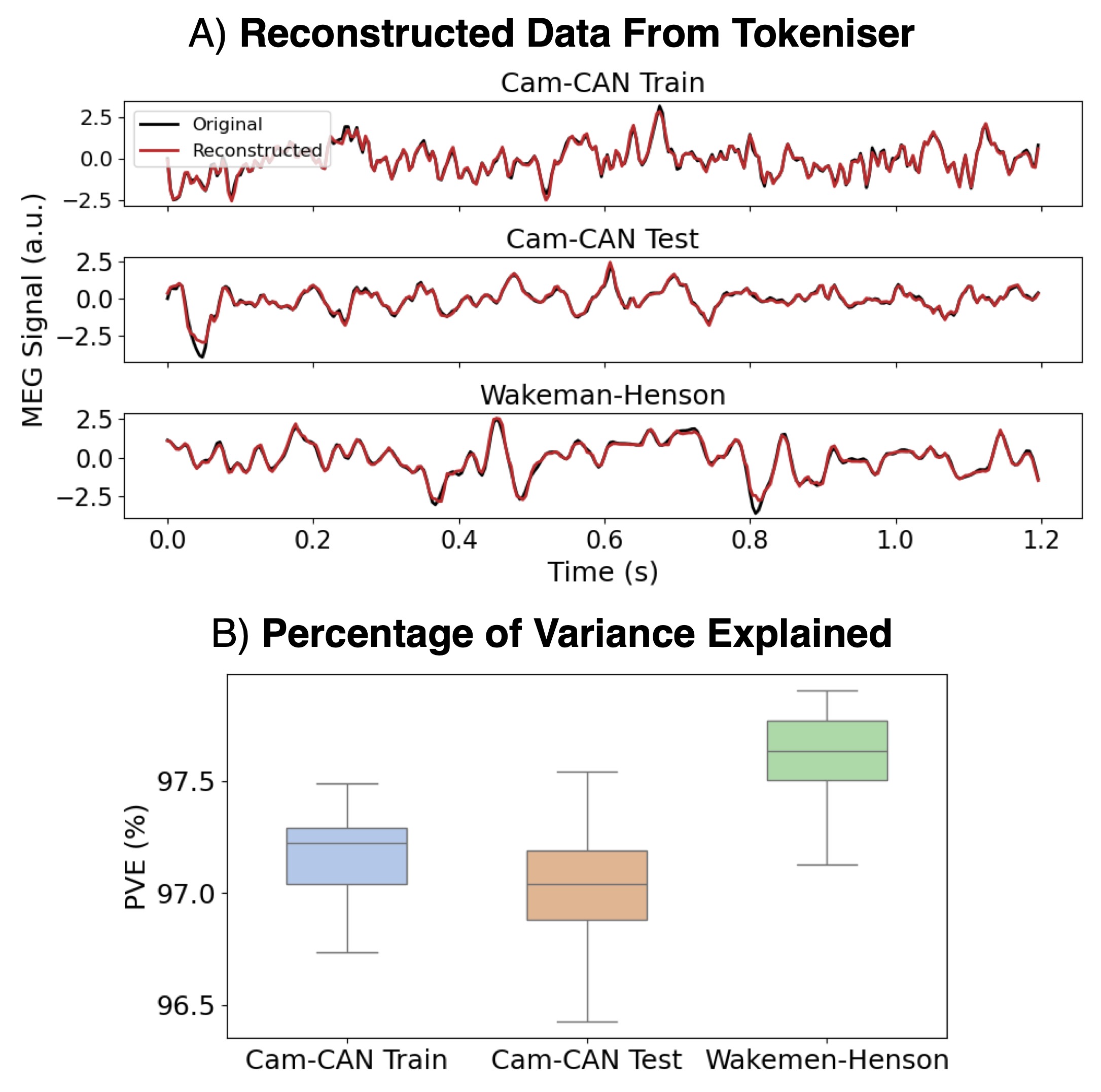}
    \caption{\textbf{The data-adaptive tokeniser reconstructs parcellated MEG data with high accuracy and generalises well to unseen data}. A) Original signals (black) and tokeniser reconstructions (where the original signals are tokenised and then de-tokenised) (red) are shown for each session from the Cam-CAN training set (top row), Cam-CAN testing set (middle row), and the Wakeman-Henson dataset (bottom row). Only the first 1.2\,s of each session are displayed. B) Percentage of variance explained (PVE) of the reconstructed data across different datasets.}
    \label{fig:tokeniser_reconstructs}
\end{figure}

Qualitatively, the tokeniser provides a high-fidelity reconstruction of the original MEG signal, closely matching its waveform across the training set, a held-out test set, and a separate, held-out dataset (Figure~\ref{fig:tokeniser_reconstructs}). To quantify this, we looked at the percentage of variance explained (PVE) of the reconstructed data using the tokeniser vs the original MEG parcel time courses (defined in Appendix~\ref{sec:extra_tokeniser_details}). We calculated the PVE on three datasets: the training set of the tokeniser in Cam-CAN (Cam-CAN train), the testing set in Cam-CAN (Cam-CAN test) and the Wakeman-Henson dataset. As shown in Figure~\ref{fig:tokeniser_reconstructs}B, with only $K^*=61$ tokens, the tokeniser achieved more than 97\% PVE on most of the sessions and also generalised to unseen data, with only a slight drop of PVE in Cam-CAN test compared with Cam-CAN train. This high reconstruction accuracy was maintained on the Wakeman-Henson dataset, confirming that the tokeniser generalises well. Notably, reconstruction performance was highest on the Wakeman-Henson dataset.

\subsection{MEG-GPT captures spatial and spectral characteristics of real data} \label{sec:results/generated_data}

We tokenised all 612 recording sessions in the Cam-CAN dataset using the trained tokeniser from Section~\ref{sec:results/tokeniser_reconstructs} and trained the \texttt{MEG-GPT} foundation model on the Cam-CAN dataset (see Section~\ref{sec:methods/training_and_testing_datasets} for details regarding the training and validation split). The subject ID (index) was also passed as an extra input to \texttt{MEG-GPT}. More details on choices of hyperparameter and training curves are available in Appendix~\ref{sec:extra_meg-gpt_details}.

After training \texttt{MEG-GPT} on the full Cam-CAN dataset, we evaluated its ability to capture the spatio-spectral features of real brain activity. We generated 60 seconds of synthetic data for each subject (see Section~\ref{sec:methods/generating_data}) and calculated the power spectral density (PSD) for every brain parcel using Welch's method (2s window, $50\%$ overlap)~\citep{welch2003use}. For a direct comparison, we generated data using a linear autoregressive (AR) model (see Appendix~\ref{sec:linear_autoregressive_model}) with the same receptive field as \texttt{MEG-GPT} to serve as a baseline.

\begin{figure}
    \centering
    \includegraphics[width=\linewidth]{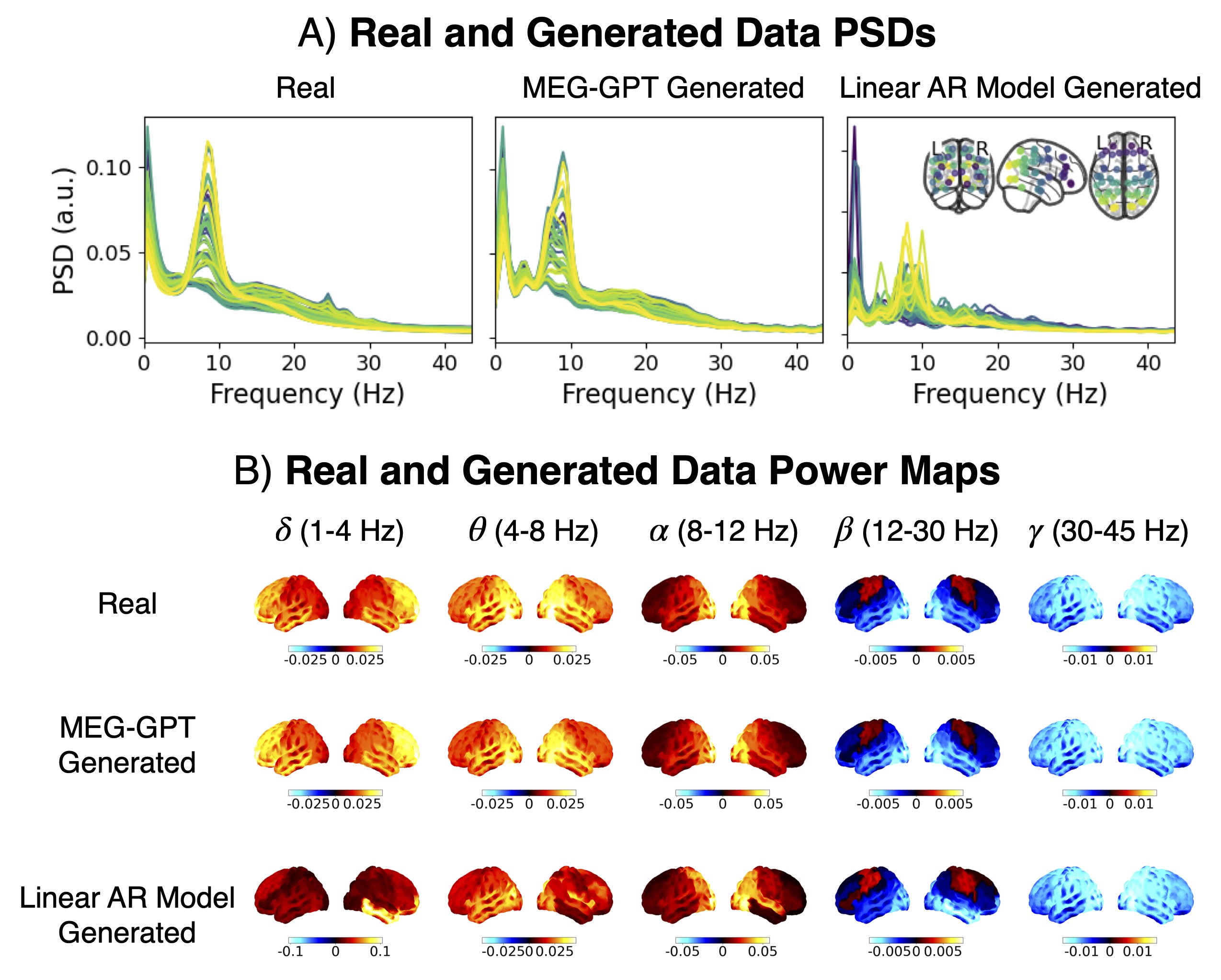}
    \caption{\textbf{MEG-GPT captures spatial and spectral characteristics of real data}. Each plot is calculated for the real data, \texttt{MEG-GPT} generated data, and linear AR model generated data. A) Group-average PSD for each parcel. The glass brain plot in the top right indicates the location of each parcel. B) Narrow-band power maps relative to the average across frequency bands.}
    \label{fig:generated_data}
\end{figure}

Figure~\ref{fig:generated_data}A shows the group-average PSD for each parcel for the generated data from both \texttt{MEG-GPT} and the linear AR model. Qualitatively, we see that \texttt{MEG-GPT} outperforms the linear AR model in capturing key features seen in the real data's PSD, e.g. the 1/f component and the size of the alpha peak. Next, in Figure~\ref{fig:generated_data}B, we integrated the PSD over five different frequency bands and plotted the spatial map of power (variance). Again, \texttt{MEG-GPT} is superior at capturing the characteristics of the real data, e.g. the frontal power in $\delta$ and $\theta$, compared to the linear AR model.

\subsection{MEG-GPT captures subject-specific fingerprints} \label{sec:results/subject_variability}

As mentioned in Section~\ref{sec:results/generated_data}, subject ID (index) were included as an extra input to \texttt{MEG-GPT} during training. We next investigated whether \texttt{MEG-GPT} learnt to generate data with subject-specific characteristics.

\begin{figure}[H]
    \centering
    \includegraphics[width=\linewidth]{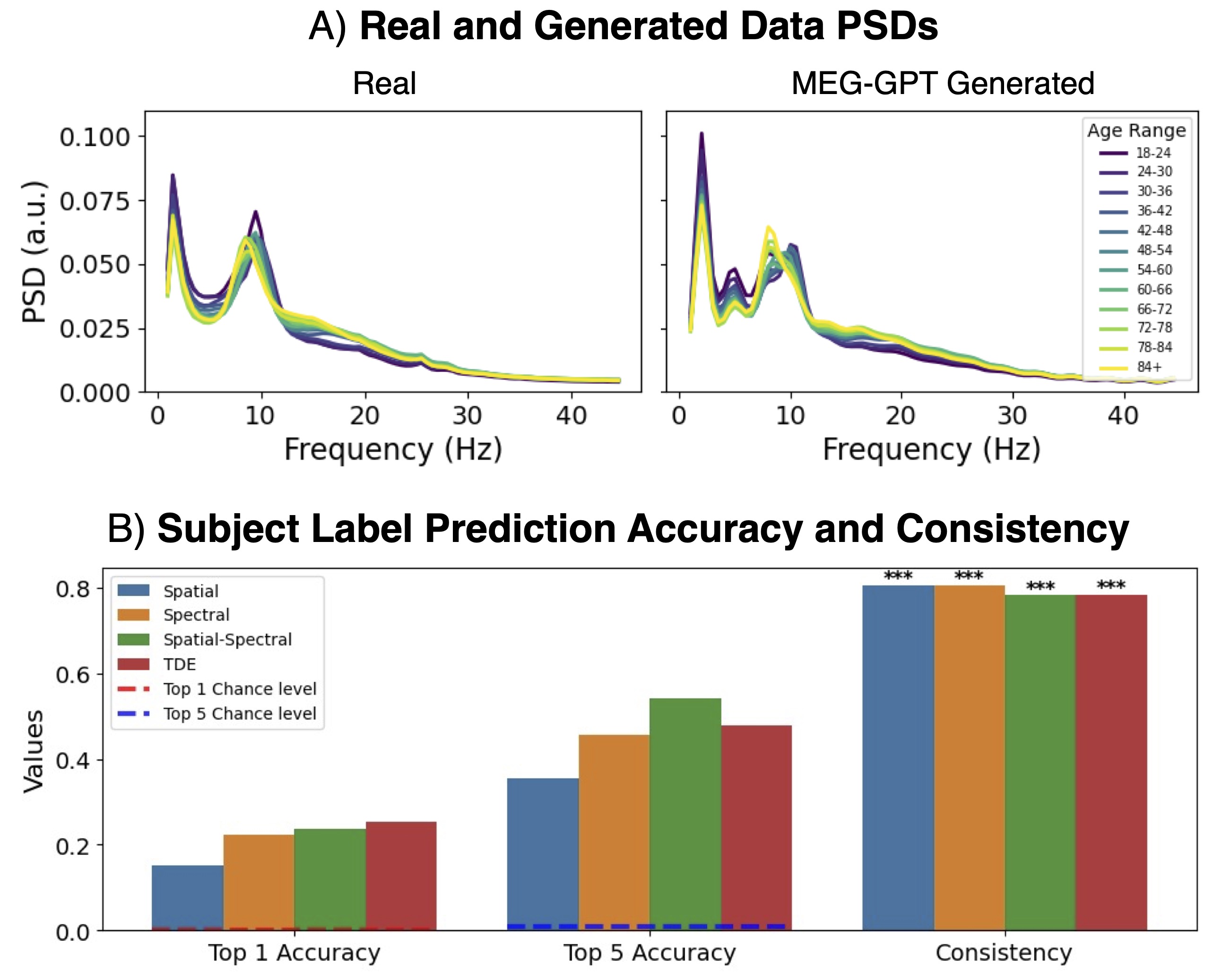}
    \caption{\textbf{MEG-GPT captures subject-specific fingerprints}. A) PSD across different age groups for real data (left) and \texttt{MEG-GPT} generated data (right). B) Top-1 accuracy, top-5 accuracy of predicting subject labels, and consistency score for 4 different features are shown. Chance level of top 1 (red dotted line) and top 5 (blue dotted line) are also illustrated. The asterisks (***) indicate a $p$-value$\,< 0.001$.
    }
    \label{fig:subject_variability}
\end{figure}

As an initial test, we examined if the model could reproduce the well-known effects of ageing on neural oscillations~\citep{quinn2024glm, gohil2024effects}. Both the real data and data generated by \texttt{MEG-GPT} successfully replicated the decrease in $\alpha$-peak frequency and increase in $\beta$ power observed in older subjects (Figure~\ref{fig:subject_variability}.

Next, we performed a more stringent test to determine if the model could generate unique individual ``fingerprints''. We extracted four different features from the real data and the \texttt{MEG-GPT} generated data (see Appendix~\ref{sec:extra_details_on_subject_variability/subject_specific_features}). For each of the features, a nearest neighbour classifier (where correlation was used as the a measure of similarity) was used for classifying subject labels. The results show that TDE features yielded the highest top-1 accuracy (Figure~\ref{fig:subject_variability}B). However, this did not persist for the top-5 accuracy where combining spatial and spectral information yielded the highest top-5 classification accuracy, confirming that \texttt{MEG-GPT} learns multifaceted fingerprints expressed in both domains.

Finally, to assess whether \texttt{MEG-GPT} has learnt the relationships between subjects, we measured the consistency score (defined in Appendix~\ref{sec:extra_details_on_subject_variability/consistency_score}). This measures the strength of agreement between the subject-pairwise distance within real data subjects and generated data subjects. All four features yielded significantly high consistency scores (under permutation test, Appendix~\ref{sec:extra_details_on_subject_variability/consistency_score}) (last column of Figure~\ref{fig:subject_variability}B).




\subsection{MEG-GPT captures bursting dynamics in MEG data} \label{sec:results/bursting_motor}

\begin{figure}[H]
    \centering
    \includegraphics[width=\linewidth]{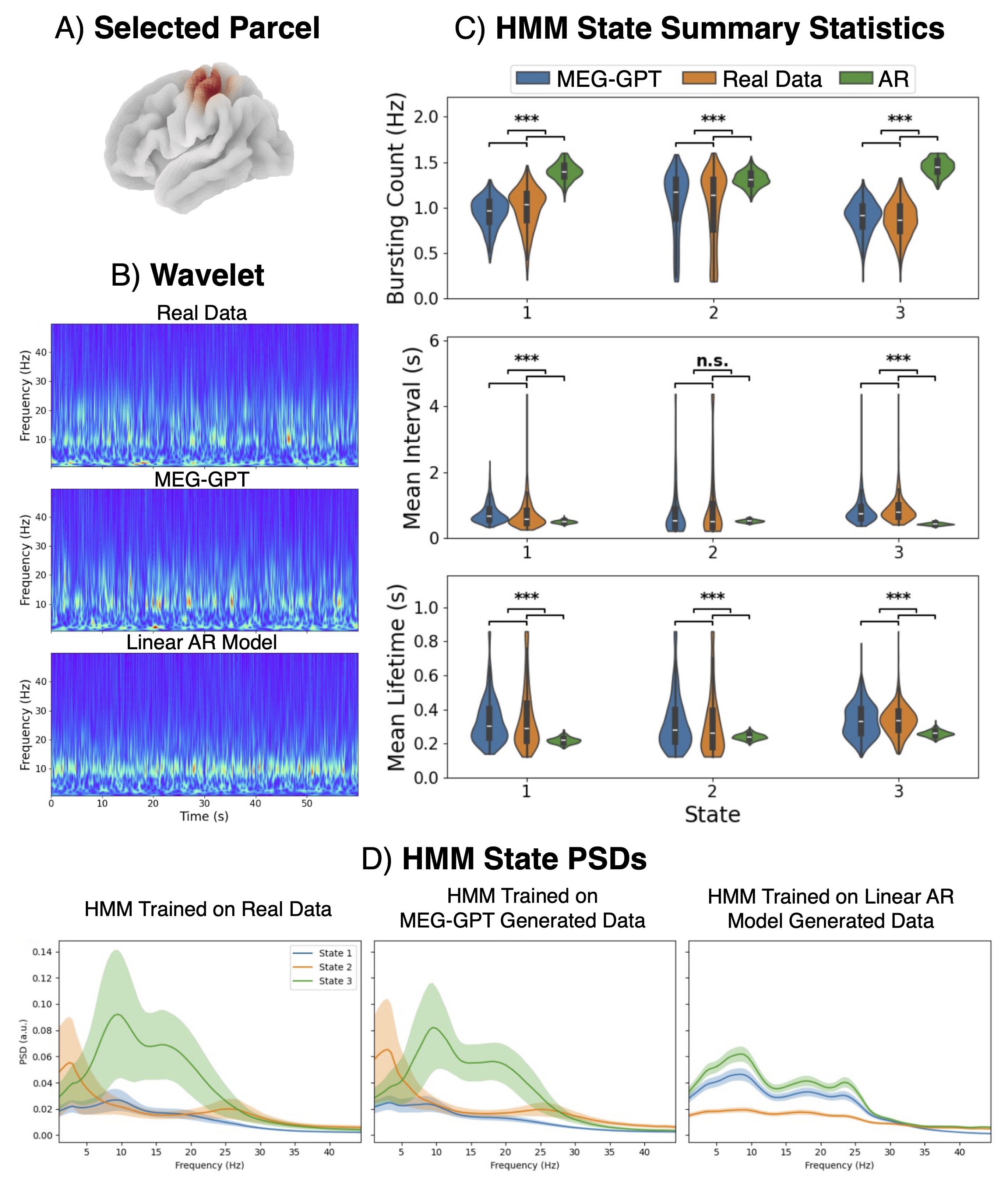}
    \caption{\textbf{MEG-GPT captures region-specific bursting dynamics in MEG data}. A) Location of the selected motor parcel. B) Wavelet transform of the first 60\,s from the first subject: real data (top), \texttt{MEG-GPT} generated data (middle), and linear AR model generated data (bottom). C) Summary statistics for each HMM state, including bursting count (top), mean interval (middle), and mean lifetime (bottom), are plotted for each of the HMM states. Results for \texttt{MEG-GPT} are shown in blue, real data in orange, and linear AR model in green. Asterisks mark statistics and states where \texttt{MEG-GPT} results more closely matched to real data compared to the AR model. The asterisks (***) indicate a $p$-value$\,< 0.001$, and ``n.s.'' indicates a non-significant result. D) State-specific PSD profiles with the solid line representing the group average and the shaded area indicating one standard deviation across subjects. The analysis was successfully reproduced using a visual cortex parcel, confirming the model's ability to capture region-specific bursting dynamics (Appendix~\ref{sec:hmm_for_burst_detection/visual}).}
    \label{fig:bursting_motor}
\end{figure}

We next investigated the ability of \texttt{MEG-GPT} to capture transient spectral bursting,  an important characteristic of brain activity that is only apparent when temporal or trial averaging is not carried out~\citep{quinn2019unpacking}. Initially, we focussed on a parcel in the motor cortex, an area known to exhibit bursting in the $\beta$ band~\citep{bonaiuto2021laminar}. We qualitatively examined the spectrograms and found that \texttt{MEG-GPT} generates data with transient bursting in the $\beta$ and $\alpha$ bands in a similar manner to the real data, whereas the linear AR model overly produces continuous bursting, particularly in the $\alpha$ band (Figure~\ref{fig:bursting_motor}.

To quantitatively characterise these bursting dynamics, we employed a Time-Delay Embedded Hidden  Markov Model (TDE-HMM) (see Appendix~\ref{sec:hmm_for_burst_detection/real_data}). The combination of TDE and HMM has been shown to be a reliable way to capture state-specific oscillatory bustings in single channel MEG data~\citep{quinn2019unpacking, gohil2024osl}, without the need to pre-specify frequency bands, amplitude thresholds, or durations of bursts. We first established a ground truth by fitting a 3-state HMM to the real data, which identified states 2 and 3 corresponding to activity in the $\delta/\theta$ and $\alpha/\beta$ bands respectively (Figure~\ref{fig:bursting_motor}D, left panel). This is discussed further in Appendix~\ref{sec:hmm_for_burst_detection/real_data}.

We then applied the same HMM inference procedure independently to both the \texttt{MEG-GPT} and linear AR model generated data \footnote{The order of states output by the HMM is arbitrary and so we used the Hungarian algorithm to match the order from different runs using the state covariance matrices (Appendix Figure~\ref{fig:burst_detection_real_data}A).}. Remarkably, we found that the \texttt{MEG-GPT} generated data uncovered a set of three states whose spectral profiles were nearly identical to those found in the real data (Figure~\ref{fig:bursting_motor}D, middle panel). In contrast, the HMM completely fails on the linear AR generated data (Figure~\ref{fig:bursting_motor}D, right panel).

Finally, we compared the temporal statistics of the HMM states across the datasets (see Appendix~\ref{sec:hmm_for_burst_detection/summary_statistics}). The summary statistics for \texttt{MEG-GPT}'s states -- including the burst count, mean interval, and mean lifetime -- were significantly more similar to the real data statistics than those from the AR model (Figure~\ref{fig:bursting_motor}C). This entire analysis was successfully repeated on a channel in the visual cortex. We found that the findings were reproduced albeit with different bursting behaviour to the motor cortex, confirming the model's ability to capture region-specific bursting dynamics (Appendix~\ref{sec:hmm_for_burst_detection/visual})

\subsection{MEG-GPT extracts features that enhance decoding performance} \label{sec:results/fine_tune}

To demonstrate \texttt{MEG-GPT}'s practical utility, we evaluated its features on a downstream visual decoding task from the Wakeman-Henson dataset. We trained a single, group-level multinomial logistic regression classifier to predict four distinct task labels (i.e. famous faces, unfamiliar faces, scrambled images, and button press). The classifier's performance was then assessed on its ability to generalise to unseen data in two challenging scenarios: new sessions from subjects seen during training (``Within Subject'') and data from an entirely new, held-out participant (''New Subject'') (subject 19 in this figure). Three feature sets were compared: \textbf{baseline} (i.e. raw epoched time-courses), \textbf{zero-shot} (i.e. epoched and time-averaged output from \texttt{MEG-GPT}'s Transformer Decoder) and \textbf{fine-tuned} (i.e. as zero-shot, but after fine-tuning the training data from Wakeman-Henson) (see Section~\ref{sec:results/fine_tune}).

\begin{figure}[H]
    \centering
    \includegraphics[width=0.9\linewidth]{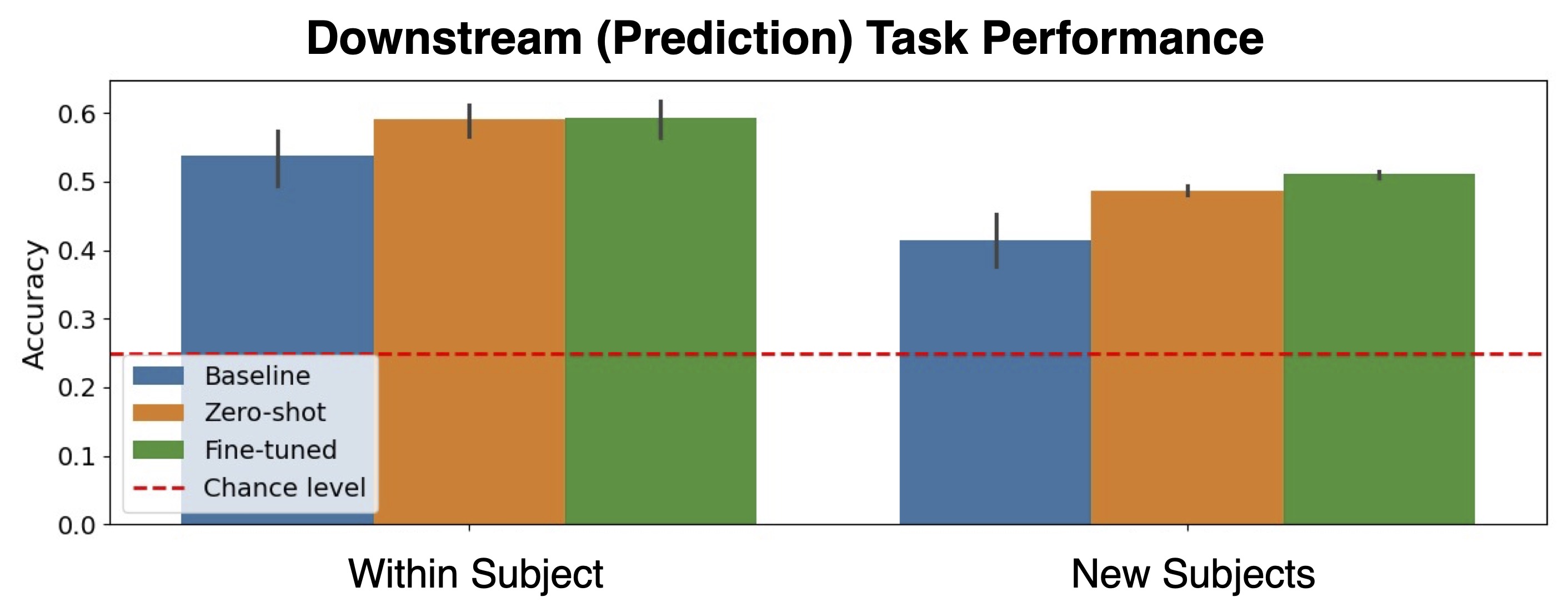}
    \caption{\textbf{MEG-GPT extracts features that enhance decoding accuracy}. Group-level multinomial logistic regression was used to predict task labels in a visual task MEG dataset (i.e. famous faces, unfamiliar faces, scrambled images, and button press). Three different feature sets are compared: baseline (i.e. raw time-courses), zero-shot (i.e. time-averaged output from \texttt{MEG-GPT}'s Transformer Decoder) and fine-tuned (i.e. same as zero-shot, but after fine-tuning on training data from Wakeman-Henson). Generalisation to unseen data was assessed ``Within Subject'' (i.e. held-out sessions from subjects already seen) [left], and to ``New Subjects'' (i.e. held-out participants) [right]. Within subject accuracies and new subject accuracies of each of the three approaches are plotted. The error bars are 95\% confidence intervals over sessions and the chance level is indicated by the red dotted line at 0.25.}
    \label{fig:decoding}
\end{figure}

As shown in Figure~\ref{fig:decoding}, features extracted from \texttt{MEG-GOT} provided a substantial boost in decoding accuracy compared to the baseline. In the \textbf{zero-shot} setting, simply using features from a pre-trained \texttt{MEG-GPT} improved ``Within Subject'' accuracy from 0.54 to 0.59 and, more dramatically, ``New Subject'' accuracy from 0.41 to 0.49. \textbf{Fine-tuning} the model on the task data provided an additional, targeted benefit for the most difficult scenario, further increasing ``New Subject'' accuracy to 0.51. These results demonstrate that \texttt{MEG-GPT} learns generalisable representations of brain activity that can significantly enhance the performance of simple linear decoders. A breakdown of performances over different labels is also demonstrated by the confusion matrices in Figure~\ref{fig:confusion_matrices} and it shows the enhancement in performance is universal across all labels.

In Figure~\ref{fig:decoding}, the ``New Subject'' results were computed using only participant 19 as the held-out subject. We repeated the analysis for baseline and zero-shot features with all other subjects being held out in turn, and the conclusion holds (Appendix~\ref{sec:extra_results_task_decoding/other_subjects}). A corresponding analysis for fine-tuned features was not possible due to the computational cost in fine-tuning the model over all train-test splits.

\section{Discussion} \label{sec:discussion}

in this work, we have demonstrated the feasibility and power of applying a self-supervised, transformer-based foundation model to continuous MEG signals. Our results show that \texttt{MEG-GPT} not only learns to generate realistic MEG data with complex, non-stationary dynamics (Figures~\ref{fig:generated_data}-\ref{fig:bursting_motor}) but also provides highly generalisable features that significantly improve decoding performance on unseen data and subjects (Figure~\ref{fig:decoding}). This work establishes a promising new framework for large-scale modelling in neuroscience, though several key areas for future development remain.

\subsection{Tokeniser}
\texttt{MEG-GPT} is designed to receive tokenised data so that it can leverage the benefits of the cross-entropy loss function. The proposed novel tokeniser is a data-driven approach that does not aim to compress the data -- this means that it does no regularisation and has no loss of temporal or spatial resolution. The philosophy behind this is that we consider the downstream transformer-based foundation model the best place to do any modelling of the rich spatio-temporal structure, rather than risk doing a relatively poor job during tokenisation. We showed that the tokeniser reconstructs MEG data with high fidelity and generalises well to unseen subjects and datasets (Section~\ref{sec:results/tokeniser_reconstructs}).

One important aspect of the tokeniser is the use of 1D convolution kernels in the decoder. The reconstruction of the continuous data based on these kernels affects both future and past time points, i.e. a token at time $t$ can invoke activity at time points before $t$. This may potentially leak information temporally when paired with a causal AR foundation model. A natural extension of the tokeniser is to implement a causal decoder.

\subsection{Modelling transient spatio-spectral structure}

\texttt{MEG-GPT} captures the spectral and spatial characteristics of MEG data, as shown in the PSD of its generated signals (Section~\ref{sec:results/generated_data}). However, the current receptive field of 80 samples (320\,ms at 250\,Hz) limits the model's ability to capture slow, low-frequency fluctuations. Extending the context window is an important future direction for improving the model.

At present, spatial information is encoded via learnable channel embeddings, which capture anatomical organisation (Appendix~\ref{sec:extra_meg-gpt_details/embeddings}), but the model lacks explicit inter-channel dependencies. This results in generated signals that are incapable of expressing functional connectivity, either static or dynamic, that is known to be present in MEG/EEG~\citep{gohil2024osl, gohil2024dynamic}. Incorporating cross-channel attention or a similar interaction mechanism is an important extension for the model.

In Section~\ref{sec:results/bursting_motor}, we demonstrated that \texttt{MEG-GPT} does an excellent job of capturing single region transient bursting dynamics, such as beta bursts in sensorimotor cortex. These features are not at all well represented by linear AR models. More broadly, the ability to inspect and validate the generated data itself (rather than evaluating the model solely through downstream performance) offers an important assessment of the fidelity of the foundation model. This approach parallels developments in generative vision and language models, where qualitative examination of generated samples has become central to understanding representational capacity. Here, we do this more quantitatively assessing the bursting dynamics of neural data. We showed that the activity generated by \texttt{MEG-GPT} reproduces hallmark neurophysiological patterns such as transient beta bursts, which are increasingly recognised as critical markers of brain function in health and disease~\citep{quinn2019unpacking, lundqvist2024beta, khanna2025separable, torrecillos2023average}.

\subsection{Learning individual differences}

Functional neuroimaging data shows a high level of variability across a population. There is increasing awareness that there is a wealth of information present in how subjects vary across population, and that large-scale datasets and more sophisticated modelling approaches are needed to unlock this potential~\citep{seghier2018interpreting}. \texttt{MEG-GPT} represents a step forward in this direction.

In \texttt{MEG-GPT}, individual variability is captured using subject/session embedding vectors. These have been successfully used in computational neuroscience to characterise individual variability~\citep{huang2024modelling, csaky2023group, chehab2021deep, defossez2023decoding, jayalath2024brain}. Section~\ref{sec:results/subject_variability} highlighted \texttt{MEG-GPT}'s ability to learn subject-specific information (fingerprints) in the data through the use of subject embeddings. This aspect of the model allows \texttt{MEG-GPT} to improve individual predictions by leveraging information from similar subjects. Further work is needed to see the impact foundation models like \texttt{MEG-GPT} can have on scientific and clinical studies, when the models are fine-tuned on bespoke studies~\citep{he2022meta}.

\subsection{Fine tuning and feature generalisability}

\texttt{MEG-GPT} demonstrates strong zero-shot performance: features extracted from the trained model on the Cam-CAN dataset generalise to new subjects and sessions in the Wakeman-Henson dataset without further training. The extracted features improve decoding performance in both within and across-subject prediction tasks (Section~\ref{sec:results/fine_tune}). This makes the model attractive for researchers with limited computational resources or those working with small datasets. We also showed that \texttt{MEG-GPT} can be efficiently fine tuned on new datasets. Pre-training on Cam-CAN took $\sim 400$ GPU hours, whereas fine tuning on the Wakeman-Henson dataset required only 3 GPU hours. This adaptability provided by fine tuning is critical for deployment across different scanner types, sampling rates, and study populations, and supports domain adaptation for diverse real-world use cases.


\subsection{Scaling to larger multi-modal datasets}

Both \texttt{MEG-GPT} and its tokeniser were trained exclusively on MEG data. A defining feature of foundation models is their scalability across diverse data modalities. MEG datasets are often limited in size and availability compared to EEG. Hence future research should explore cross-modal training on both MEG and EEG. This raises challenges such as handling differences in sampling rate, channel layout, and number of sensors (which could be circumvented by data preprocessing, source reconstruction and parcellation), but offers a compelling opportunity to investigate whether \texttt{MEG-GPT} can capture the shared and modality-specific structure of brain activity across recording techniques.

\subsection{Limitations and future directions}

In this work, when using \texttt{MEG-GPT} derived features on downstream decoding task, we simply averaged the features over time to reduce dimensionality for decoding tasks. While this yields strong performance, it potentially discards temporal dynamics encoded in the token sequence. This may disadvantage \texttt{MEG-GPT} relative to the baseline approach in some settings. Developing classification architectures that retain temporally resolved representations, e.g., via convolutional, recurrent or attention-based decoders, could further improve task performance.

Our decoding experiments use a linear classifier trained on extracted features. Prior work~\citep{yuan2024brainwave} suggests that end-to-end fine-tuning of both the decoder and a more flexible classification head yields superior results. Future work should explore fine tuning the entire model to directly optimise supervised objectives.

\section{Conclusion} \label{sec:conclusion}

In this work, we introduced \texttt{MEG-GPT}, a foundation model trained on large-scale resting-state MEG data, and a novel, data-driven tokeniser that is used to provide \texttt{MEG-GPT} with tokenised data inputs. Our results demonstrate that this self-supervised approach is highly effective: the model generates realistic data capturing complex neural dynamics, and its features significantly improve zero-shot generalisation in downstream decoding tasks. This confirms the feasibility of applying large-scale generative models to electrophysiological signals, paving the way for versatile, powerful, and general-purpose tools in neural decoding and computational neuroscience.

\section*{Ethics statement}

The study that collected the CamCAN dataset was conducted in compliance with the Helsinki Declaration, and had been approved by the local ethics committee, Cambridgeshire 2 Research Ethics Committee. Written informed consent was given by participants. See \citep{shafto2014cambridge} for details regarding protocols. The Wakemen-Henson dataset (\citep{wakeman2015multi}) was approved by Cambridge University Psychological Ethics Committee. Written informed consent was obtained from participants.

\section*{Data and code availability statement}

Data used are publicly available. For the Wakeman-Henson dataset, we refer the readers to the original paper \citep{wakeman2015multi}. For the Cam-CAN dataset, we refer the readers to the original paper \citep{taylor2017cambridge}.

Source code and scripts for reproducing results in the paper are available on GitHub: \url{https://github.com/OHBA-analysis/osl-foundation}. Example code and tutorials for training a foundation model and applying it to new data are also provided.

\section*{Credit authorship contribution statement}

RH: Conceptualisation, Methodology, Software, Validation, Formal analysis, Investigation, Data curation, Writing - original draft, Writing - review and editing, Visualisation. SC: Conceptualisation, Methodology, Software, Writing - review and editing, CG: Conceptualisation, Data curation, Writing - original draft, Writing - review and editing. MW: Conceptualisation, Methodology, Data curation, Writing - review and editing, Supervision.

\section*{Acknowledgements}
Research is supported by the Wellcome Trust (106183/Z/14/Z, 215573/Z/19/Z), the New Therapeutics in Alzheimer’s Diseases (NTAD) study supported by UK MRC, the Dementia Platform UK (RG94383/RG89702) and supported by the NIHR Oxford Health Biomedical Research Centre (NIHR203316). The views expressed are those of the author(s) and not necessarily those of the NIHR or the Department of Health and Social Care. The Wellcome Centre for Integrative Neuroimaging is supported by core funding from the Wellcome Trust (203139/Z/16/Z and 203139/A/16/Z). RH is supported by the Wellcome Trust (215573/Z/19/Z). SC is supported by the Medical Sciences Graduate School Studentship, funded by the Medical Research Council (MR/W006731/1), the Hertford Claire Clifford Lusardi Scholarship, and the Nuffield Department of Clinical Neurosciences. CG is supported by the Wellcome Trust (215573/Z/19/Z). OPJ is supported by the MRC (MR/X00757X/1), Royal Society (RG\textbackslash R1\textbackslash 241267), NSF (2314493), NFRF (NFRFT-2022-00241), and SSHRC (895-2023-1022). MW is supported by the Wellcome Trust (106183/Z/14/Z, 215573/Z/19/Z).

\section*{Declaration of competing interest}

No competing interests.

\bibliographystyle{apalike}
\bibliography{references}

\clearpage

%
%

\setcounter{figure}{0}
\counterwithin{figure}{section}
\counterwithin{table}{section}

\setcounter{equation}{0}
\renewcommand\theequation{\thesection.\arabic{equation}}

\section*{Supplementary Information (SI)}

\appendix

\section{Extra details regarding the tokeniser} \label{sec:extra_tokeniser_details}

\subsubsection*{Percentage of variance explained}

The percentage of variance explained (PVE) is defined by
\begin{equation}
    PVE = 100 \times \left(1 - \frac{\sum_{t=1}^T \sum_{c=1}^C (x_t^{(c)} - \tilde{x}_t^{(c)})^2}{\sum_{t=1}^T \sum_{c=1}^C (x_t^{(c)})^2} \right) \%,
\end{equation}
where $x$ is the real data, $\tilde{x}$ is the reconstructed data, $T$ is the number of time points and $C$ is the number of channels.

\subsubsection*{Hyper-parameters and training curve}

\begin{figure}[H]
    \centering
    \includegraphics[width=0.75\linewidth]{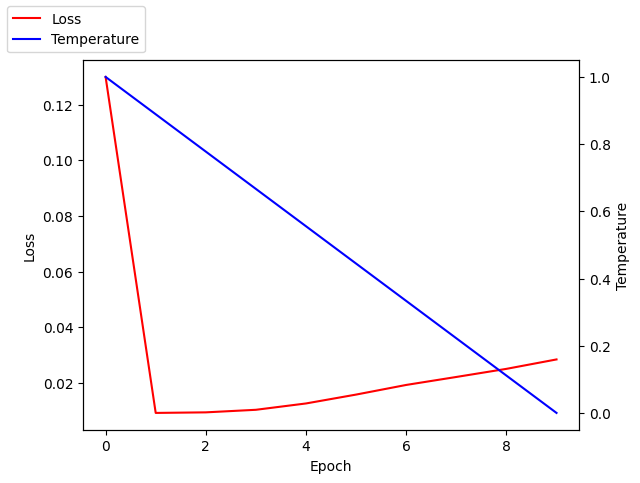}
    \caption{\textbf{Training curve of the tokeniser}. The training loss is plotted in red against the epochs and the temperature during annealing is plotted in blue.}
    \label{fig:tokeniser_loss}
\end{figure}

Here we present the hyper-parameters of the tokeniser in Table \ref{tab:tokeniser_hyperparameters} and the training curve in Figure \ref{fig:tokeniser_loss}.

\begin{table}[H]
    \centering
    \begin{tabular}{|cc|cc|}\hline
         \multicolumn{4}{|c|}{\textbf{Model parameters}}\\\hline
         Number of tokens $K$&  128&  Token width $d_{token}$&  10\\
         GRU number of units&  128&  GRU sequence length &  200\\ \hline
 \multicolumn{4}{|c|}{\textbf{Training parameters}}\\\hline
         Batch size&  32&  Number of epochs&  10\\
 Learning rate& 1e-5& &\\ \hline
    \end{tabular}
    \caption{Hyper-parameters for the tokeniser.}
    \label{tab:tokeniser_hyperparameters}
\end{table}

\begin{figure}[H]
    \centering
    \includegraphics[width=\linewidth]{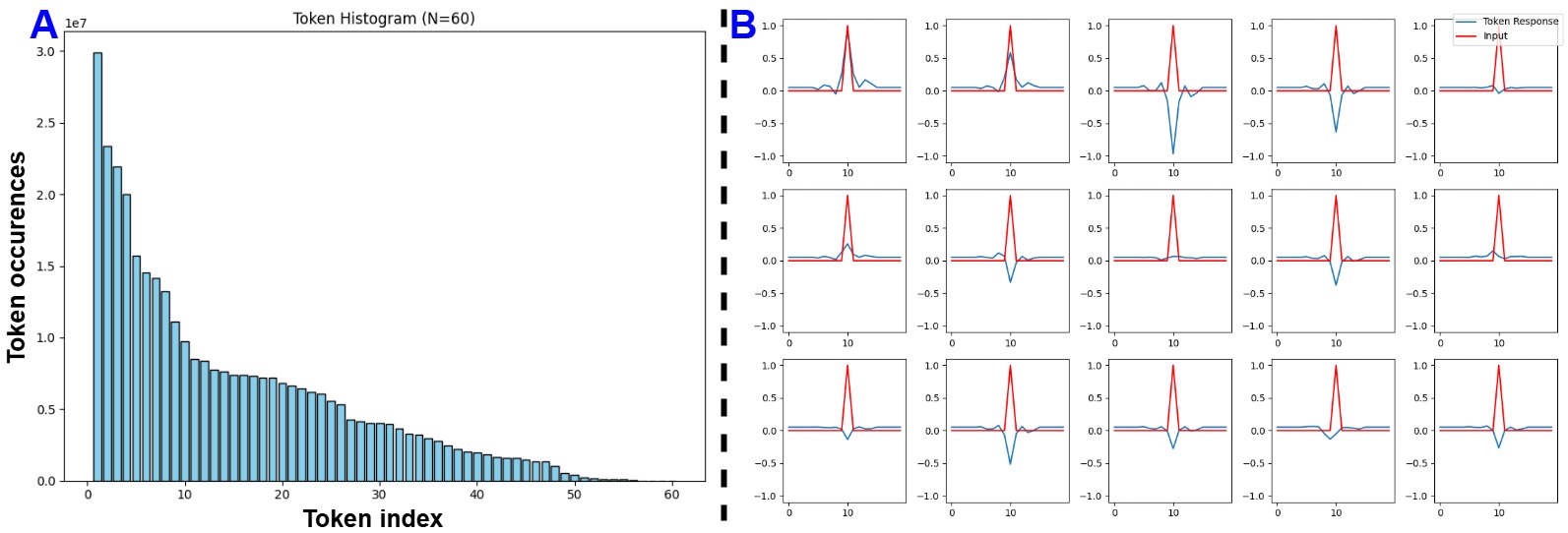}
    \caption{\textbf{Summary plots of the learnt tokens}. (A) Histogram of token occurrences. (B) Shape of top 15 token response kernels learnt during training. Here we show the output (blue) of convolving the token kernels with a pulse (red).}
    \label{fig:token_summary}
\end{figure}

\section{Extra details regarding MEG-GPT} \label{sec:extra_meg-gpt_details}

\subsection{Model hyper-parameters} \label{sec:extra_meg-gpt_details/hyperparameters}

Here we present details of the hyper-parameters used for training MEG-GPT on the Cam-CAN dataset (Table \ref{tab:foundation_hyperparameters_camcan}) and on the Wakeman-Henson dataset (Table \ref{tab:foundation_hyperparameters_wh}).

\begin{table}[H]
    \centering
    \begin{tabular}{|cc|cc|}\hline
         \multicolumn{4}{|c|}{\textbf{Model parameters}}\\\hline
         Token embedding dimension $d_z$&  400&  Channel embedding dimension $d_c$&  400\\
         Position embedding dimension $d_p$&  400&  Subject ID embedding dimension $d_s$&  400\\
 Input embedding dimension& 400& Transformer model dimension $d$&400\\
 Receptive field $L$& 80& Patch size $L_p$&4\\
 Number of patches $P$& 20& Unpatched sequence length $L_u$&16\\
 Latent sequence length $L_{latent}$ & 40& Number of head $N_{head}$&4\\
 Number of layers $N_{layer}$& 4& Feed forward network number of units&400\\
 Feed forward network activation& Leaky ReLU & Feedforward network dropout&0.2\\\hline
 \multicolumn{4}{|c|}{\textbf{Training parameters}}\\\hline
         Batch size&  8&  Number of epochs&  60\\
 Learning rate& 1e-5& Loss sequence length $L_{loss}$&8\\ \hline
    \end{tabular}
    \caption{\textbf{Hyper-parameters for the foundation model on Cam-CAN}.}
    \label{tab:foundation_hyperparameters_camcan}
\end{table}

\begin{table}[H]
    \centering
    \begin{tabular}{| c |c |}
    \hline
    Training parameter & Value \\
    \hline
         Batch size &  16 \\
         Number of epochs & 10\\
         Learning rate & 5e-7\\
    \hline
    \end{tabular}
    \caption{\textbf{Hyper-parameters for fine-tuning the foundation model on Wakeman-Henson}.}
    \label{tab:foundation_hyperparameters_wh}
\end{table}

\subsection{Training curves} \label{sec:extra_meg-gpt_details/training_curves}

\begin{figure}[H]
    \centering
    \includegraphics[width=0.48\linewidth]{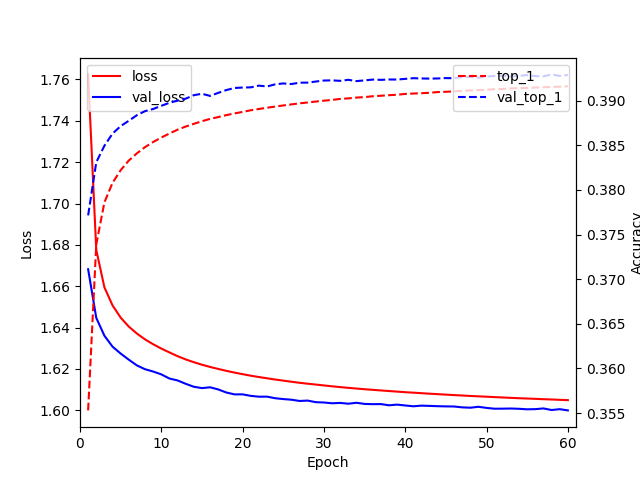}
    \includegraphics[width=0.48\linewidth]{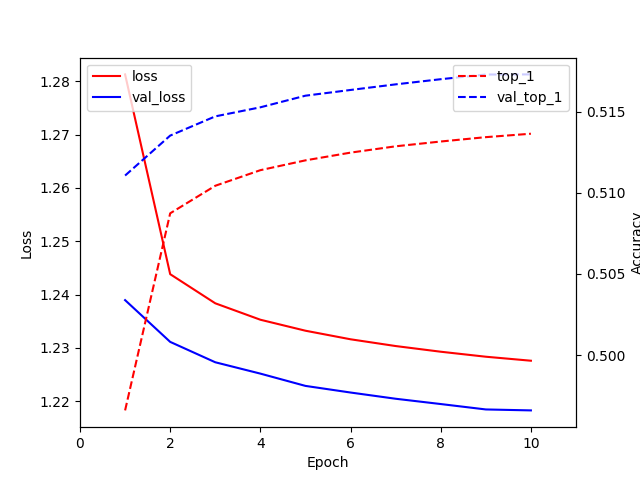}
    \caption{\textbf{Training curves of MEG-GPT}. Left: Training curve on the Cam-CAN dataset. Right: Training curve on the Wakeman-Henson dataset.}
    \label{fig:meg-gpt_history}
\end{figure}

Here we present the training curves of MEG-GPT when trained on the Cam-CAN dataset and fine-tuned on the Wakeman-Henson dataset in Figure \ref{fig:meg-gpt_history}.

\subsection{Embedding vectors encode meaningful information through training} \label{sec:extra_meg-gpt_details/embeddings}

\begin{figure}[H]
    \centering
    \includegraphics[width=\linewidth]{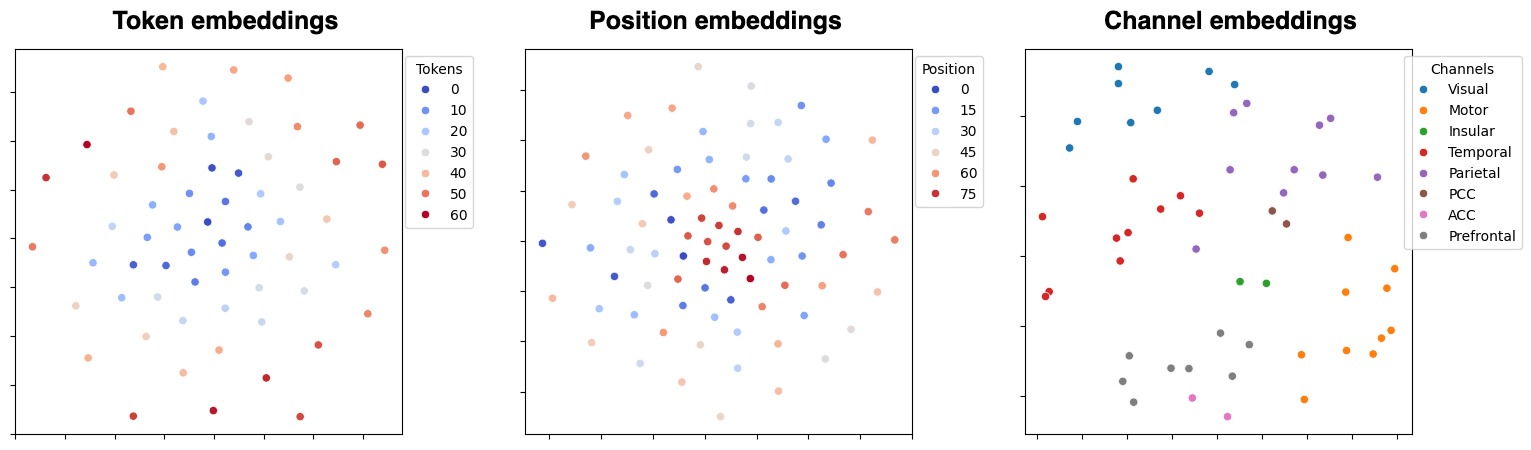}
    \caption{\textbf{Embedding vectors encode meaningful information}. Each of the embeddings are projected with tSNE to 2 components.}
    \label{fig:embeddings}
\end{figure}

Here we visualise the different embedding vectors learnt during the training process, with tSNE, shown in Figure \ref{fig:embeddings}. We can see that token embeddings and position embeddings are organised in their respective embedding spaces according to token frequency and position in a sequence. Furthermore, we see well-separated clusters of the channel embeddings according to pre-defined cortical regions, and regions that are geographically close are also close in the channel embedding space (e.g. Visual channels are close to temporal and parietal channels). This shows different sources of meaningful variations in the data are captured by MEG-GPT during the training process.

\section{Extra details regarding subject variability} \label{sec:extra_details_on_subject_variability}

\subsection{Subject-specific features} \label{sec:extra_details_on_subject_variability/subject_specific_features}

Here we describe four different ways to extract subject-specific features.

\textbf{Spatial feature}: This feature is designed to only include spatial information and no spectral information. For MEG data of each subject, we first calculate the PSD of the data and take the average of the PSD over frequencies for each channel.

\textbf{Spectral feature}: This feature is designed to only include spectral information and no spatial information. For MEG data of each subject, we first calculate the PSD of the data and take the average of the PSD over channels for each frequency bin.

\textbf{Spatial + Spectral feature}: This feature is designed to include both spatial and spectral information. For MEG data of each subject, we first calculate the PSD of the data and flatten the matrix to a vector.

\textbf{Time delay embedding feature}: This feature is also designed to include both spatial and spectral information. It performs the time-delayed embedding transformation \citep{vidaurre2016spectrally} by adding lagged versions of each channel as extra channels of the data. The static covariance matrix of the time-delay embedded data is calculated for each subject and the upper triangular part of the covariance matrix is flattened to give the required feature.

\subsection{Predicting subject labels from features} \label{sec:extra_details_on_subject_variability/predict_subject_labels}

Here we describe how we predict subject label of MEG-GPT generated data based on features introduced in Appendix \ref{sec:extra_details_on_subject_variability/subject_specific_features}. Conceptually, for each subject $i$ we check if feature of subject $i$ in the generated data is closest to the feature of subject $i$ in the real data compared to that of all other real data subjects. More formally, let $x$ and $\hat{x}$ be the real and generated data. We further let $f$ and $\hat{f}$ be the extracted features (any of the features introduced in Appendix~\ref{sec:extra_details_on_subject_variability/subject_specific_features}) from $x$ and $\hat{x}$, respectively. Then we use a nearest neighbour classifier with correlation distance as the distance metric. More specifically, we construct the pairwise correlation distance matrix $\Sigma_{x, \tilde{x}} \in \mathbbm{R}^{N_{subject} \times N_{subject}}$ such that the $i, j$-th entry of $\Sigma$ is the correlation distance ($1 - $ correlation) between subject $i$ in the real data and subject $j$ in the generated data. To calculate the top k accuracy, we check for each column of $\Sigma_{x, \tilde{x}}$ if the diagonal element is among the smallest $k$ elements in the column.

\subsection{Consistency score} \label{sec:extra_details_on_subject_variability/consistency_score}

Consistency score is a measure of the similarity between the pairwise structure in real data and generated data. We start with computing correlation matrices $\Sigma_{x,x}$ and $\Sigma_{\tilde{x}, \tilde{x}}$, which are pairwise correlation matrices of the features between real data subjects and between generated data subjects respectively. Then the consistency score is defined as the correlation between the upper triangular elements of $\Sigma_{x,x}$ and $\Sigma_{\tilde{x}, \tilde{x}}$.

Given the null distribution that generated data of different subjects is the same, the features extracted have the same distribution and is exchangeable. Hence we can permute the rows and columns of $\Sigma_{\tilde{x}, \tilde{x}}$ and get the null distribution of the consistency score. 

\section{Linear autoregressive model} \label{sec:linear_autoregressive_model}

In this paper we also trained linear autoregressive models as a baseline comparison with MEG-GPT. To account for the fact that different channels might have different data distributions, we train independent autoregressive models on data of each of the channels. The linear autoregressive model used in this paper has an order of 80 in order to match the receptive field of MEG-GPT.

During data generation of each channel, an initial prompt is generated from a standard normal distribution. Then data are generated in an autoregressive manner where at each time point, Gaussian noise with standard deviation being the standard error of the regression is used.

\section{Hidden Markov Model for burst detection} \label{sec:hmm_for_burst_detection}

\subsection{Extra results of TDE-HMM on real data} \label{sec:hmm_for_burst_detection/real_data}

\begin{figure}[H]
    \centering
    \includegraphics[width=\linewidth]{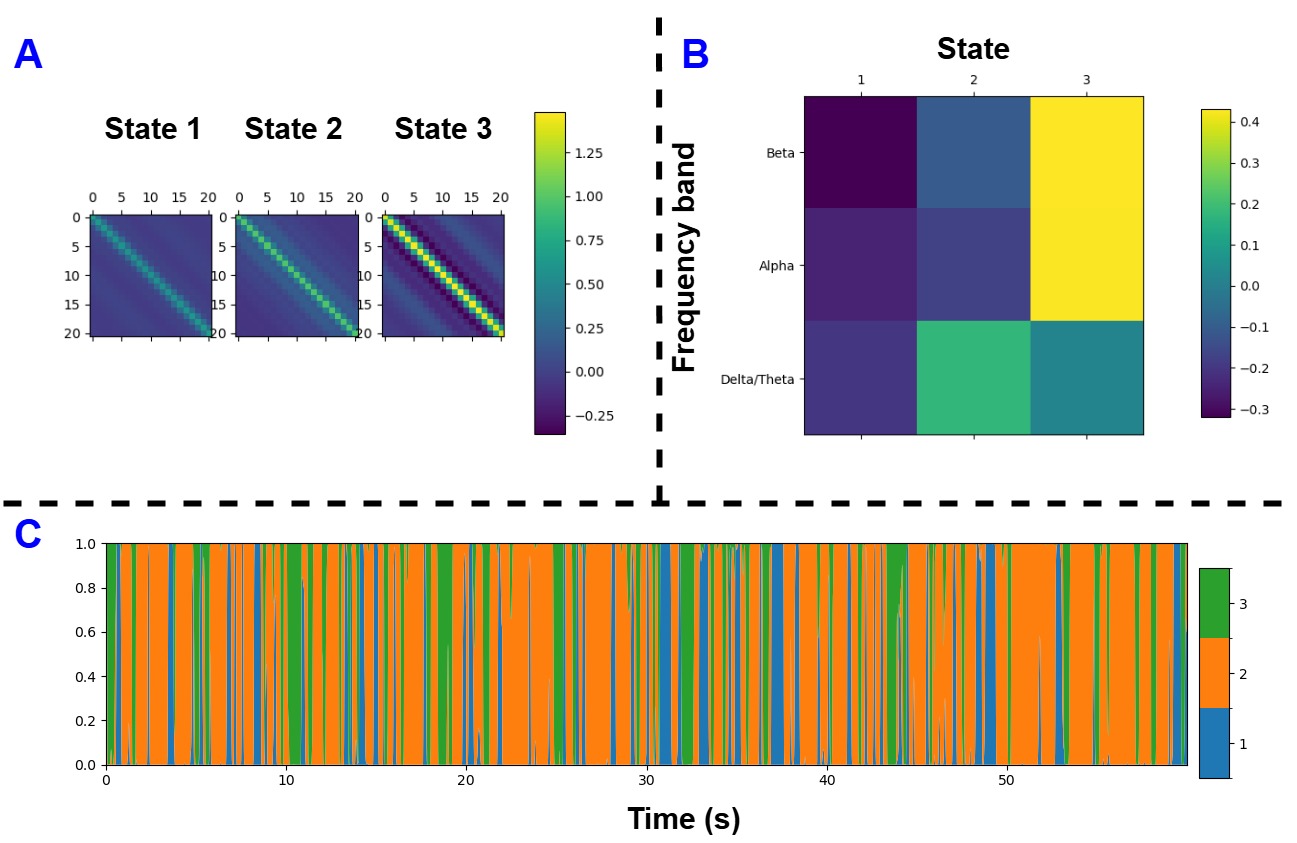}
    \caption{\textbf{Results of TDE-HMM applied to real MEG data}. (A) Inferred state covariance matrices for the three hidden states. (B) Correlation between state time courses and power in different frequency bands. (C) Exmaple state time courses over the first 60 seconds from the first subject in the Cam-CAN dataset.}
    \label{fig:burst_detection_real_data}
\end{figure}

Here we show extra results of single channel burst detection on real data. We can see from Figure \ref{fig:burst_detection_real_data}B that state 3 is correlated with increased $\alpha$ and $\beta$ power and state 2 is correlated with increase $\delta/\theta$ power. In addition, we present the state covariance matrices and inferred state time courses in Figure \ref{fig:burst_detection_real_data}A and \ref{fig:burst_detection_real_data}C.

\subsection{Summary statistics of HMM states} \label{sec:hmm_for_burst_detection/summary_statistics}

3 summary statistics of the state time courses are used in this paper. These are
\begin{itemize}
    \item \textbf{Bursting count}: Also referred to as the switching rate, it is defined as the mean number of activations per second for each subject.
    \item \textbf{Mean interval}: It is defined as the mean time between state activations for each subject.
    \item \textbf{Mean lifetime}: It is defined as the mean time between entering and leaving a state for each subject.
\end{itemize}

\subsection{Results on a visual channel} \label{sec:hmm_for_burst_detection/visual}

\begin{figure}[H]
    \centering
    \includegraphics[width=\linewidth]{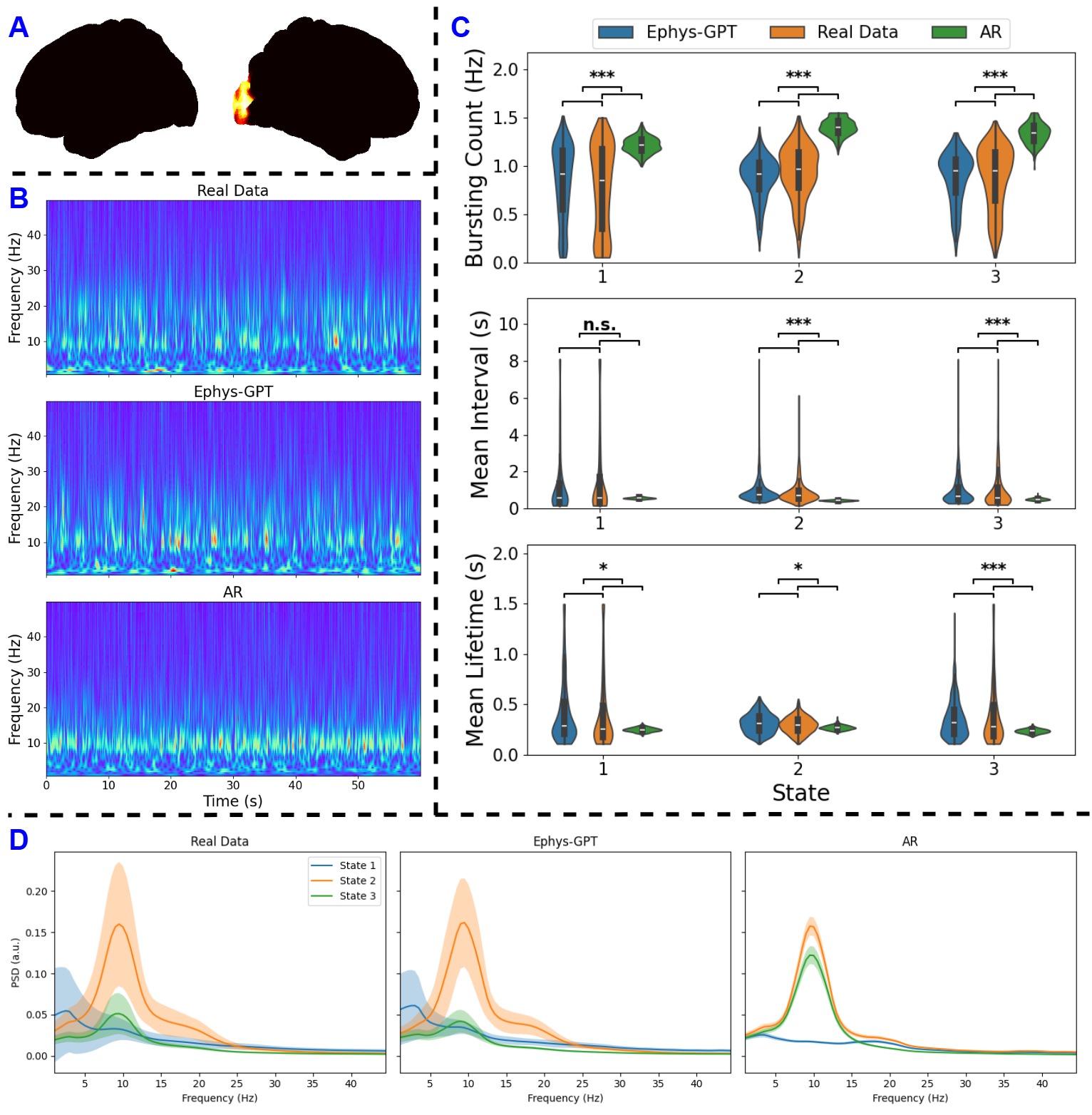}
    \caption{\textbf{Burst detection in visual cortex}. (A) Location of the visual parcel selected. (B) Wavelet transform of the first 60 seconds from the first     subject: real data (top), MEG-GPT generated data (middle), and linear autoregressive (AR) model generated data (bottom). (C) Summary statistics for each HMM state, including bursting count (top), mean interval (middle), and mean lifetime (bottom). Results for MEG-GPT are shown in blue, real data in orange, and AR model in green. Asterisks mark statistics and states where MEG-GPT results more closely match real data compared to the AR model. The asterisks (*) and (***) indicate a $p$-value$\,< 0.05$ and $< 0.001$, respectively, and ``n.s.'' indicates a non-significant result. (D) State-specific power spectral density (PSD) profiles with the solid line representing the group average and the shaded area indicating one standard deviation across subjects.}
    \label{fig:bursting_visual}
\end{figure}

Here we show that the results in Section \ref{sec:results/bursting_motor} can be reproduced in another channel. Here we choose a visual channel, location shown in Figure \ref{fig:bursting_visual}A.

\section{Extra details regarding the task decoding} \label{sec:extra_results_task_decoding}

\subsection{Classifier} \label{sec:extra_results_task_decoding/classifier}

The pipelines for classification for raw epoch features and MEG-GPT extracted features are the same. Each feature is standardised with the mean and standard deviation computed using the trials in the training set. Then a multinomial logistic regression with default settings in scikit-learn v1.7.0 is used.

\subsection{Other subjects as testing subject} \label{sec:extra_results_task_decoding/other_subjects}

In the main text, we have used subject 19 in the Wakeman-Henson dataset to test for out of subject accuracy. Here we repeated the analysis for the baseline and the zero-shot features using other subjects to test for out of subject accuracy. The results are shown in Figure \ref{fig:extra_decoding_results}A, and we see that the improvements in both within subject and out of subject accuracy are consistent over the different testing subjects.

\subsection{Session-specific classifier} \label{sec:extra_results_task_decoding/session_specific_classifier}

\begin{figure}[H]
    \centering
    \includegraphics[width=\linewidth]{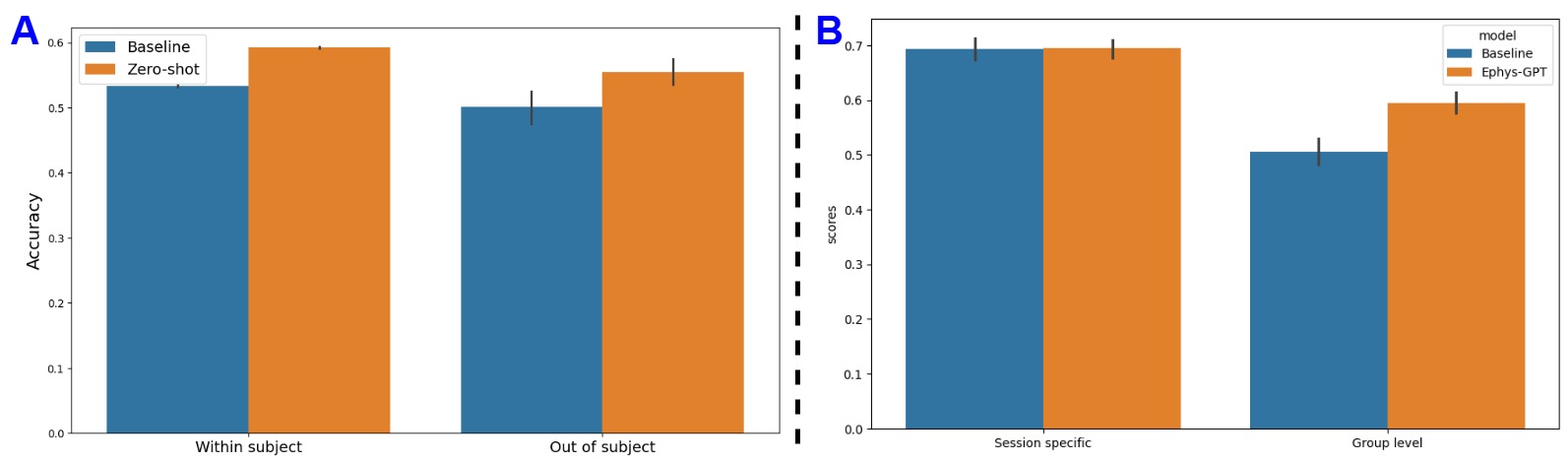}
    \caption{\textbf{Extra decoding results}. (A) Decoding accuracy when different subjects are chosen to test for within and out of subject accuracy. Here the error bars are 95\% confidence interval across subjects. (B) Decoding accuracy of session-specific and group level classifiers trained on baseline and fine-tuned MEG-GPT features. Here, the error bars are 95\% confidence interval across sessions.}
    \label{fig:extra_decoding_results}
\end{figure}

\begin{figure}[H]
    \centering
    \includegraphics[width=\linewidth]{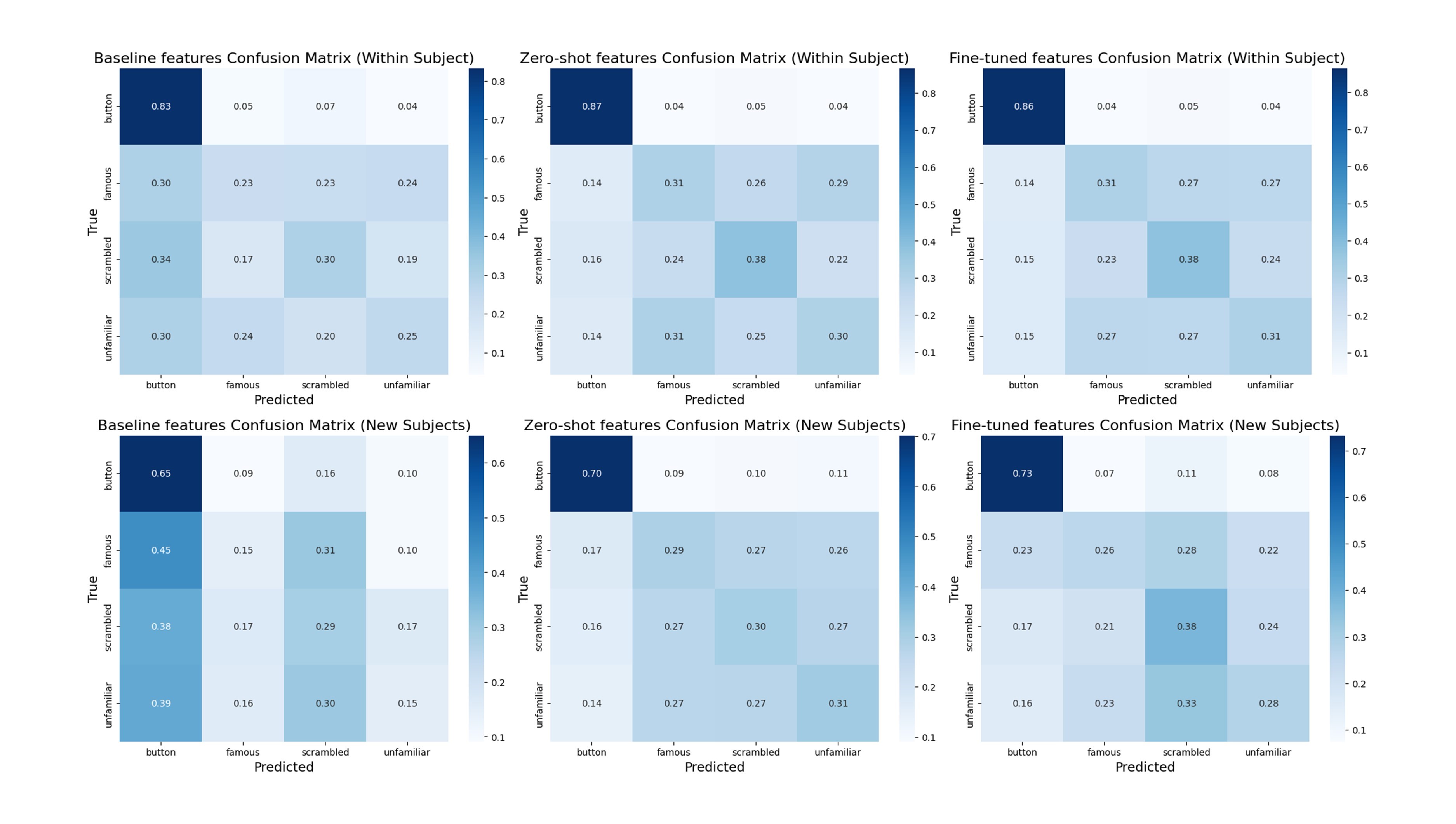}
    \caption{\textbf{Confusion matrices}. Confusion matrices of Within Subject prediction (top row) and for New Subject prediction (bottom row) are plotted for baseline features (left column), zero-shot features (middle column), and fine-tuned features (right column).}
    \label{fig:confusion_matrices}
\end{figure}

Due to the large heterogeneity across sessions, it is common to train session specific classifiers in decoding tasks, although we cannot apply the trained classifiers to new sessions and subjects. Here for each session in the Wakeman-Henson dataset, we take 90\% of the recordings as training set and 10\% as the testing set. We fine-tuned the MEG-GPT foundation model with new session embeddings learnt and extracted features from the outputs of the decoder in the same way as in Section \ref{sec:methods/feature_extractor}. In Figure \ref{fig:extra_decoding_results}B, we can see that the session-specific classifiers from baseline and fine-tuned MEG-GPT features achieved comparable accuracies while MEG-GPT features still demonstrated superior accuracy of group level classifiers. The confusion matrices are demonstrated in Figure \ref{fig:confusion_matrices}.

\end{document}